\documentclass[10pt,twocolumn,letterpaper]{article}

\usepackage[pagenumbers]{cvpr}      %

\usepackage{booktabs}
\usepackage{colortbl}
\definecolor{refgrey}{gray}{0.92}
\definecolor{camblue}{HTML}{0072B2}
\definecolor{suborange}{HTML}{E69F00}
\definecolor{memgreen}{HTML}{009E73}
\definecolor{failred}{HTML}{D55E00}
\graphicspath{{pix/}}

\definecolor{cvprblue}{rgb}{0.21,0.49,0.74}
\usepackage[pagebackref,breaklinks,colorlinks,allcolors=cvprblue]{hyperref}

\title{Hierarchy in Motion: Continuous Control of Discrete Autoregressive Video}

\author{Tharindu Wickremasinghe$^{1}$, Zeeshan Nadir$^{2}$, Harshana Weligampola$^{1}$,  Yu Yuan$^{1}$, Stanley H. Chan$^{1}$ \\
$^1$Purdue University
$^2$Samsung Research America}

\newcommand{\projecturl}{\url{https://tharindu-nirmal.github.io/HierarchyInMotion/}}
\newcommand{\projectpage}{\href{https://tharindu-nirmal.github.io/HierarchyInMotion/}{project page}}
\begin{document}
\addtocontents{toc}{\protect\setcounter{tocdepth}{-1}}

\twocolumn[{%
\renewcommand\twocolumn[1][]{#1}%
\maketitle
\vspace{-2.0em}
\begin{center}
  \centering
  \captionsetup{type=figure}
  \includegraphics[width=\textwidth]{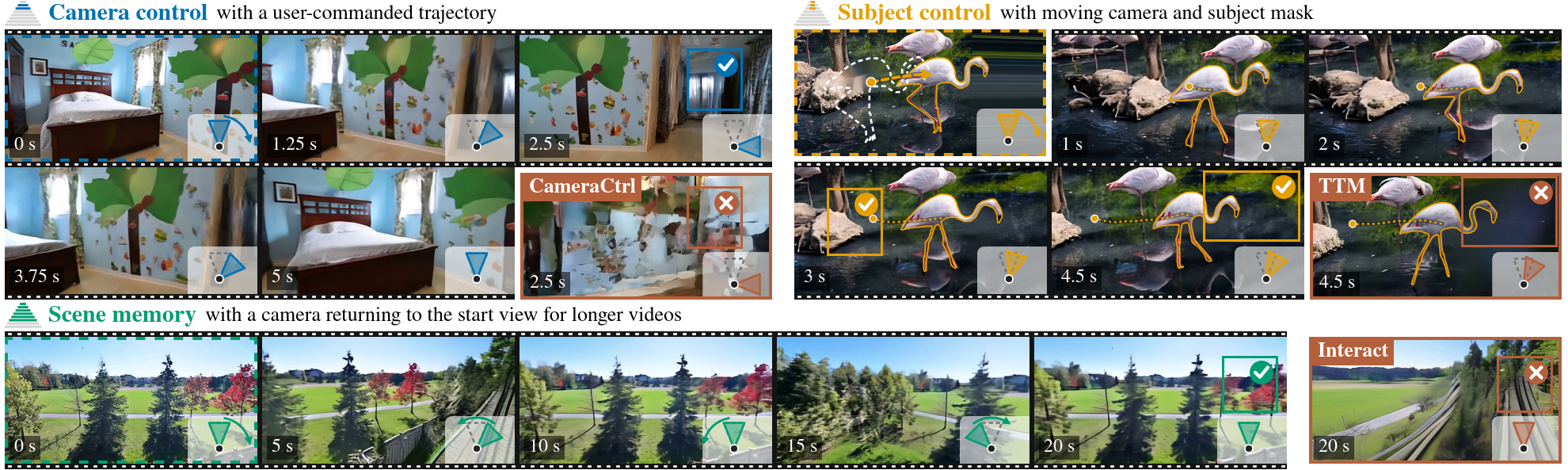}
  \captionof{figure}{\textbf{One token pyramid steers camera, subject and memory.}
  \emph{Camera:} a commanded $90^\circ$ pan right and back from one start frame.
  \emph{Subject:} a cut-and-dragged flamingo~\cite{davis2017} moves while the camera turns.
  \emph{Memory:} after 20\,s of turning right, back, left and back, our token store restores the start view.
  Pyramids: the levels that steer each task (schematic);
  brown frames: baselines at the same time.}
  \label{fig:teaser}
\end{center}%
\vspace{0.5em}
}]
\begin{abstract}
Discrete autoregressive video models generate fast, streamable clips,
and their coarse tokens can anchor a compact memory.
Yet it is unclear which tokens continuous controls such as camera geometry should steer.
Existing methods discretize controls, or condition every token without asking
where the control acts.
We find that the structure of the token hierarchy is both the source of the control challenge 
and the key to solving it.
The discrete tokens span a feature pyramid, abstracting features of scene identity at the top, 
all the way down to pixel-level detail.
Isolating which band of levels decides or renders a property gives a recipe for
continuous control: make sure the control signal reaches the band that decides it,
and retain the coarse levels as reusable scene conditioning.
We demonstrate how this recipe can guide different forms of efficient control from the
same autoregressive model. First, we steer continuous camera motion, beating comparable diffusion
models across user commanded motion types and real camera paths, in less time. 
Secondly, we move a subject along a path while a camera moves, beating its continuous 
diffusion counterpart in median path accuracy while generating $15\times$ faster.
Finally, we generate long videos efficiently with a coarse-token cache $11.6\times$  smaller than the 
full token pyramid, while revisiting scenes as accurately as models trained for memory.
All data and code will be made public at \projecturl.
\end{abstract}

\section{Introduction}
\label{sec:intro}

Video generators are becoming interactive, streamable world models~\cite{genie,cosmos,matrixgame2,gamengen}.
For this role, discrete autoregressive models offer three properties that continuous diffusion
models lack.
They inherit the machinery of language models: a single next-token objective, explicit
probabilities and key--value caching~\cite{videopoet,emu3}.
Their tokens are a codec: a clip becomes a fixed budget of bits that is cheap to store and
copy~\cite{magvitv2,bsq}.
And with next-scale prediction~\cite{var,infinity}, which writes each clip coarse to fine as a
token pyramid, they are fast: InfinityStar~\cite{infinitystar}
generates 720p video about $10\times$ faster than leading diffusion models.

Interaction, however, is continuous: moving a camera, directing a subject or returning to a
place is specified by angles, distances and trajectories, and a text prompt can ask for
``a slow pan to the left'' but not for a $45^\circ$ pan over $5$ seconds.
Continuous diffusion models control video with camera embeddings, point tracks and
masks~\cite{cameractrl,motionprompting,throughthemask}, and the most precise methods reconstruct the
scene in 3D and render it~\cite{gen3c,das}, which is too slow for interactive streaming.

\textbf{Discrete autoregressive models, although fast, are harder to control, and the difficulty is structural.}
A control signal is continuous, but each generation step is a discrete choice among
$2^{16}$ or more codes per token. A command can only shift the
probabilities of independently predicted bits, and a sampled token is never revised.
Discrete world models therefore discretize the control too, as keyboard tokens and binned
camera angles~\cite{mineworld} or learned latent actions~\cite{genie},
losing magnitude, speed and arbitrary trajectories, which works but leaves open 
\emph{where the control actually steers motion}.

We argue that the answer lies in the structure that makes these models fast.
A next-scale prediction model autoregressively predicts the next scale, forming a token pyramid.
For a video frame, this forms a hierarchy. A single token at the top, all the way down to tokens at very high resolution, 
and each level only refines what the earlier levels committed to.
What each level \emph{renders and decides}, has not been measured, and so the question that matters for
control has not been asked:
\emph{Where in a discrete coarse-to-fine video hierarchy is motion decided and rendered?
How can continuous control use it?}

We answer this by establishing a causal hierarchy on a 14-level bitwise next-scale video model (\cref{fig:teaser}).
We separate two kinds of bands within the discrete token pyramid.
\emph{Render bands} mark when a property becomes visible as levels are added: Scene, Motion, Layout and Detail.
\emph{Control bands} mark where a property is decided: the smallest set of levels whose tokens,
copied from one generation into another, carry the property with them. Through our `token transplant
experiments', we find that motion is decided in a narrow control band, before it is rendered.
Camera control on these levels alone, $0.7\%$ of the tokens, recovers $98\%$ of the gain of control at every
level, while control where motion becomes visible does nothing.

This leads to our motion control recipe.
\textbf{The key to controlling a discrete coarse-to-fine video model is to exploit its token hierarchy.
The control signal must reach the coarse levels that hold the control band. Control placed only where
motion becomes visible fails.}
Additionally, the coarse levels hold the scene's committed state as exact discrete tokens, and creates a compact memory.

We show that this recipe works for two types of motion.
First, we control a camera trajectory of static scenes with Pl\"ucker-ray conditioning~\cite{lfn}
as the continuous control. Our recipe enables the discrete autoregressive model to outperform
comparable continuous diffusion models. We show this across standard camera test data,
and user-commanded camera motions while running faster inference.
Second, we move a subject along a commanded path with a \emph{region-token transplant}. 
With a moving camera, the subject follows its guided path
and we outperform the continuous diffusion counterpart~\cite{ttm} while generating $15\times$ faster.
Finally, we use a cache of the coarse tokens, $11.6\times$ smaller than the full token pyramid. 
On long 20s video, we achieve $+10.2$\,dB over a no memory baseline and $+3.3$\,dB 
over a continuous latent store of similar size.
Combined with our camera control, we are able to steer a camera view to revisit
scenes with no separate learned memory module, as accurately as models trained for memory.

In summary, our contributions are:
\begin{itemize}
  \item \textbf{A causal map of a discrete token hierarchy.}
  We show that motion is decided at the coarse levels before it is rendered,
  through causal interventions on the generation.
  \item \textbf{A recipe for continuous control of discrete tokens.}
  The control signal must reach the coarse levels of the control band. This motivates the
  camera adapter, and designs a subject motion control and a memory reader.
  \item \textbf{Camera control, subject control, and a compact scene memory} from one
  discrete token model, outperforming their diffusion counterparts in camera control and subject control
  with a moving camera.
\end{itemize}

\section{Related Work}
\label{sec:related}

\begin{figure*}[t]
  \centering
  \includegraphics[width=\linewidth]{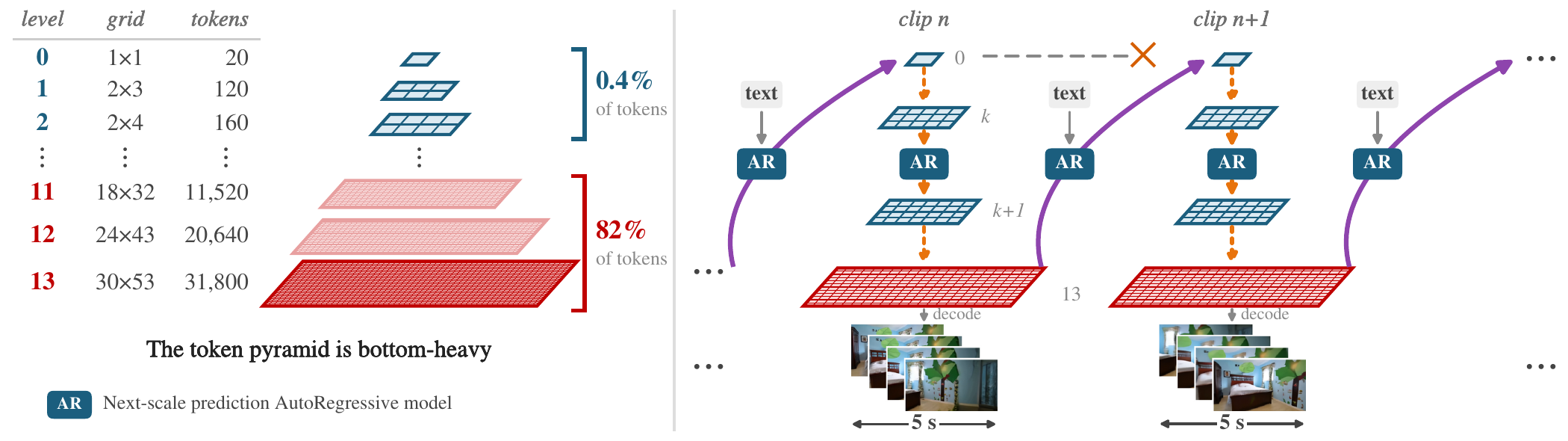}
  \caption{\textbf{Discrete next-scale video generation.}
  Left: a 5\,s clip is a bottom-heavy pyramid of 14 token maps, from a $1{\times}1$ grid per latent frame at
  level 0 to $30{\times}53$ at level 13.
  Right: Next-scale prediction with an Autoregressive transformer block. 
  Clips are generated one after another; every level of the next clip attends to 
  the finest level of the previous one (purple), and the text prompt conditions each clip.
  The coarse features are forgotten at the next clip(\textcolor{failred}{$\times$}), 
  and it is not known at which level a continuous camera signal should enter.}
  \label{fig:nextscale_combined}
\end{figure*}

\paragraph{Long, interactive video models}
generate successive clips conditioned on preceding frames~\cite{selfforcing,diffusionforcing,streamingt2v} and take keyboard
actions~\cite{matrixgame2}, learned latent actions~\cite{genie,lapo,lapa} or camera commands~\cite{cosmos}.
They need low latency, continuous control, and scene memory beyond the immediate context.

\paragraph{Discrete token video models}
match the low latency requirement for interactive video: next-scale prediction~\cite{var,infinitystar}
writes a clip in a few dozen steps. They compress each clip with a quantized autoencoder~\cite{vqvae,vqgan,
magvitv2,bsq} and model the resulting codes with a transformer, either one token at a
time~\cite{videopoet,emu3,llamagen,ivideogpt}, by masked parallel decoding~\cite{maskgit}, or one
resolution level at a time~\cite{var,infinity,infinitystar}; good video tokenizers also
compress comparably to standard codecs~\cite{magvitv2}.
We add a further advantage: because a next-scale model writes a
clip coarse to fine, its token pyramid has an inherent \emph{measurable hierarchy}, which
tells us where control must act and how stored tokens must be reused.

\paragraph{Continuous control of video generation} is commonly done with a camera control signal using
continuous-latent diffusion models. They encode the camera
as per-pixel Pl\"ucker rays and inject them through an adapter~\cite{cameractrl,ac3d};
subject motion is controlled with point tracks~\cite{motionprompting} or moving
masks~\cite{throughthemask}, or without training, by guiding attention features along box
trajectories~\cite{sgi2v} or by denoising a cut-and-dragged puppet video~\cite{ttm}.
Our coarse-token transplant also starts from such a puppet, but writes it only into the few
coarse levels that our motion control recipe identifies.
The most precise camera methods reconstruct the first frame in 3D and render it along the
target path, then let the generator repair the rendering~\cite{gen3c,viewcrafter,das,seeu}.
These methods follow trajectories well but pay for reconstruction and rendering at every call.
Camera geometry has also been placed in attention as relative encodings between
views~\cite{gta,prope}; we instead add absolute Pl\"ucker rays~\cite{lfn} as offsets to the queries and
keys of a parallel branch, and ask where in a discrete hierarchy they act.

\paragraph{Control of discrete autoregressive video.}
Discrete world models usually discretize the control as well: keyboard tokens and binned
camera angles~\cite{mineworld} or learned latent actions~\cite{genie}; even continuous
diffusion world models adopt a small vocabulary of camera primitives~\cite{disco}.
This preserves the discrete interface but loses magnitude, speed and arbitrary trajectories.
RAYNOVA~\cite{raynova} conditions a next-scale driving model on rays,
but does not measure how accurately the commanded camera is followed, or at which levels the
conditioning acts.
To our knowledge, we give the first causal map of where motion is decided in a discrete next-scale video hierarchy.
This enables to go beyond just camera control, to subject control
and a memory reader.

\paragraph{Memory for long video.}
Long-video models such as InfinityStar-Interact~\cite{infinitystar} forget the scene once it leaves
their context window: a camera that turns away and back finds a different room.
World models with memory retrieve past frames by pose~\cite{worldmem}, keep an explicit 3D
point-cloud memory~\cite{spatia,gen3c}, or warp positional encodings so that one
mechanism serves both camera control and memory~\cite{ucm}.
These memories are large, continuous, or require reconstruction.
Our memory keeps the coarse levels of the discrete pyramid itself, 88\,KB per clip.
Our token hierarchy motivates a token decoding - reordering - 
encoding design that serve as an efficient memory.

\section{Discrete Next-Scale Video}
\label{sec:prelim}

\begin{figure*}[t]
  \centering
  \includegraphics[width=\linewidth]{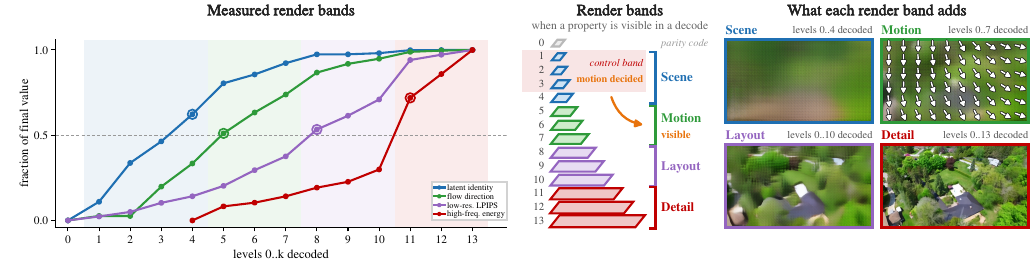}
  \caption{\textbf{Render bands,} measured on the tokenizer, mark when a property becomes visible
  on partial decode. Left: The fraction of each property's final values: Latent scene identity, Flow direction, 
  low-res. LPIPS, and high-freq. energy are tracked when only levels $0..k$ are decoded.
  Middle: the resulting bands: Scene, Motion, Layout and Detail on the token pyramid.
  Motion is decided in the \textbf{control band} (levels 1--3, measured by token transplant), but
  becomes visible only in the later Motion band. Right: one held-out clip decoded.}
  \label{fig:bands_combined}
\end{figure*}

\paragraph{The token pyramid and next-scale prediction.}
A 5\,s clip is represented as a pyramid of 14 maps of binary tokens, predicted coarse to fine
by a transformer~\cite{var,bsq,infinity,infinitystar} (\cref{fig:nextscale_combined}).
Each level refines, but never revises the coarser levels, and all tokens within a level are predicted in parallel.
The pyramid is bottom-heavy: the three finest levels contain 82\% of its tokens, 
with 64 bits per token against 16 at the coarser levels.
Training minimizes a weighted bitwise binary cross-entropy (BCE),
\begin{equation}
  \mathcal{L} = \sum_{k} \lambda_k \sum_{i\in r_k}\sum_{b}
  \mathrm{BCE}\big(\hat p_{k,i,b},\, r_{k,i,b}\big),
  \label{eq:loss}
\end{equation}
where $r_{k,i,b}$ is bit $b$ of token $i$ at level $k$ and $\lambda_k$ is its level's weight;
given the context, the bits are predicted independently.
The fine levels add appearance and detail; the hierarchy of what each level renders is shown
in \cref{fig:bands_combined}.

\paragraph{Clip-by-clip generation.}
Long videos are generated one clip at a time: each new clip attends to the finest level of the
previous clip through a key--value cache (\cref{fig:nextscale_combined}, right).
This makes generation streamable and lets the user change the command between clips, but gives
each clip only one clip of scene context.
A view that remains out of frame for multiple clips may therefore be forgotten on return, while
the finest level is also the most costly to cache.

\paragraph{Two open questions.}
\label{sec:prelim:gap}\label{sec:prelim:hard}
A command can only shift the probabilities of independently predicted bits, and a sampled level
is never revised, so it must act before the levels that settle the property; storing all levels
is costly.
\emph{Which levels should a continuous signal steer, and which should be kept as memory?}
\Cref{sec:method} answers both.

\section{The Hierarchy Is the Key}
\label{sec:method}

We measure what each level does in two ways (\cref{sec:method:render,sec:method:control}) and
turn the measurement into a recipe (\cref{sec:method:recipe}).
A \emph{render band} is a range of levels in which a property becomes \emph{visible} when decoded.
The \emph{control band} is the shortest prefix of levels $0..k$ whose tokens, copied into another
generation, carry the property with them, named by the levels that add transfer.
\Cref{fig:bands_combined} (middle) shows both maps; they do not coincide.

\subsection{Render bands: a property becomes visible}
\label{sec:method:render}

We encode 80 held-out RealEstate10K~\cite{re10k} clips, decode them up to each level, and compare
four properties of each partial decode with the full decode (\cref{fig:bands_combined}, left).
A band begins where its property first passes $50\%$ of its final value, giving four bands in a
fixed order: Scene (1--4), Motion (5--7), Layout (8--10) and Detail (11--13) (\cref{sec:supp_bands}).
Content from partial pyramids can be out of distribution for the decoder, so render bands mark
when a property becomes visible in a decode, not when the tokens contain it.
Therefore, conditioning motion only where it becomes visible can fail, and the next experiment shows
why. 

\subsection{Control bands: a property is decided}
\label{sec:method:control}

\begin{figure}[!t]
  \centering
  \includegraphics[width=\linewidth]{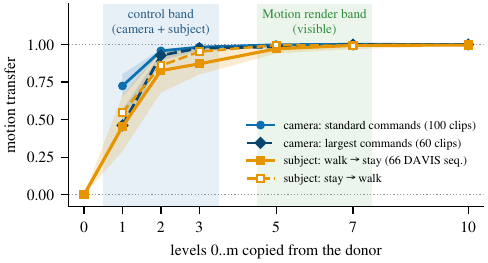}
  \caption{\textbf{Token transplant:} Motion transfer against the number 
  of levels copied from a donor generation. 
  Lines: median over clips; shading: 95\% bootstrap confidence interval of the median.
  Both types of motion, camera(blue) and subject(yellow) are decided in 
  the control band (levels 1-3: blue shading), before the motion render band (green).}
  \label{fig:transfer}
\end{figure}

\paragraph{Token transplant.}
To find where a property is decided we need an intervention, not a correlation.
We generate two clips from the same first frame, caption and seed, so that they
differ only through a command.
A \emph{donor} under command $A$ and a \emph{recipient} under command $B$.
We then re-run the recipient with its levels $0..m$ forced to the donor's tokens and let the
model generate the rest under command $B$.
If the property of the output follows the donor, it was decided in levels $0..m$.
\Cref{fig:transfer} reports \emph{motion transfer}: the output's mean optical flow projected 
onto the donor-minus-recipient flow, so that 0 means ``moves like the recipient'' and 1 ``moves like the donor''.

\paragraph{Motion is decided at levels 1--3.}
We test camera motion in two settings: standard motions on held-out RealEstate10K~\cite{re10k}
scenes, and larger user-guided commands such as $90^\circ$ pans and tilts.
Standard motions transfer with levels 0--2, and the largest commands also use level 3
(\cref{fig:transfer}).
These levels override even the command: a call told to pan left, given the coarse tokens of a
right pan, pans right, although its frames still look clearly different from the donor's.
Motion is therefore fixed before appearance.
Subject motion behaves the same way: most of it transfers by level 2 and nearly all by level 3.
Since both controls share these levels, camera and subject must be separated spatially rather
than by level. Scenes, models, per-command numbers, and a decoder free logit probe that agrees
with this is in \cref{sec:supp_control}.

\subsection{The recipe: control that reaches the band, memory in the coarse levels}
\label{sec:method:recipe}

\begin{figure}[t]
  \centering
  \includegraphics[width=\linewidth]{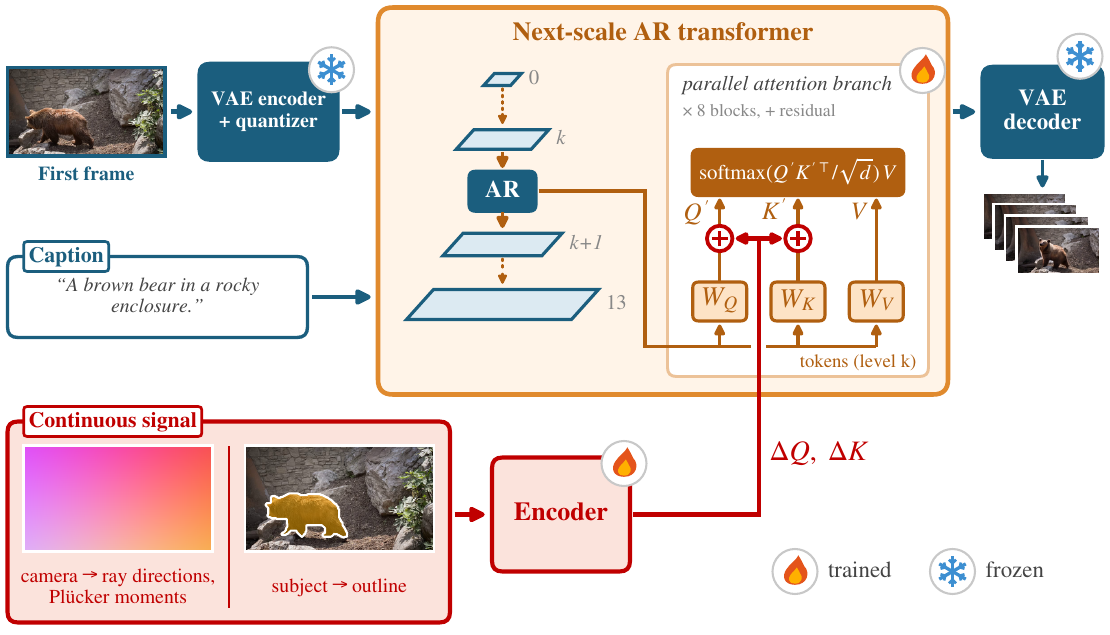}
  \caption{\textbf{The training recipe.}
  A continuous signal (camera rays, or a subject's outline) is encoded into $\Delta Q,\Delta K$
  of a trained attention branch that runs parallel to 8 blocks and adds to their residual stream.}
  \label{fig:arch}
\end{figure}

\begin{figure}[t]
  \centering
  \includegraphics[width=\linewidth]{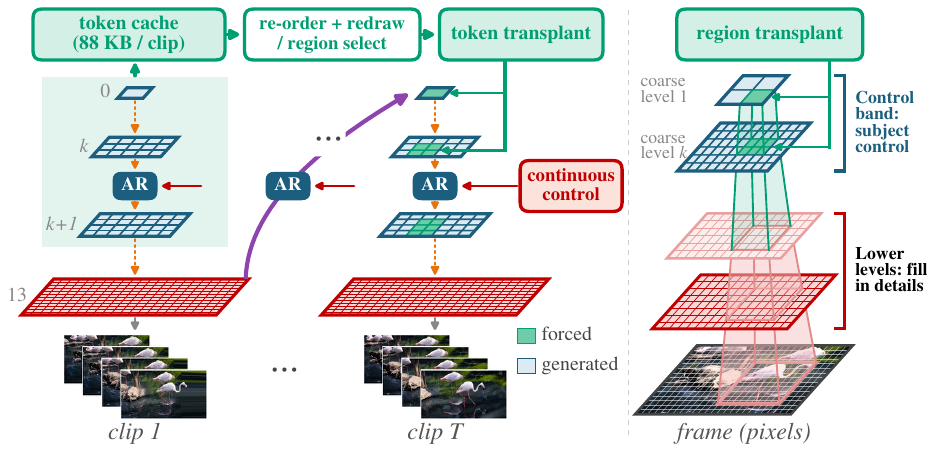}
  \caption{\textbf{Inference recipe: force tokens at the coarse levels.}
  Left: each clip's levels 0--10 are cached; a later clip is forced with them, re-ordered to its
  camera path with the detail redrawn, or with the subject region of a puppet video. The model of
  \cref{fig:arch} generates the rest under the continuous control.
  Right: a region transplant forces only the subject's tokens in the control band.}
  \label{fig:redraw}
\end{figure}

\begin{figure*}[!t]
  \centering
  \includegraphics[width=\linewidth]{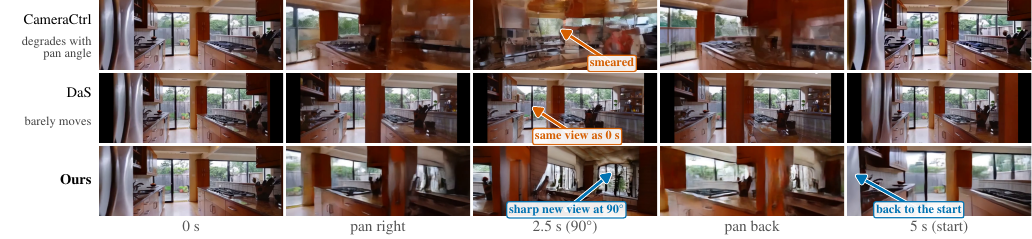}%
  \caption{\textbf{Pan right $90^\circ$ and back, same first frame and command.}
  Held-out RealEstate10K~\cite{re10k} start frame.
  Our model, (with camera control on all levels) turns closer to the commanded $90^\circ$ with a sharp new view and returns to the start.
  DaS barely moves, and CameraCtrl smears as the pan angle grows. We encourage readers to refer the videos on the \projectpage{} and \cref{sec:supp_camera_visual}.}
  \label{fig:camera}
\end{figure*}

\paragraph{1. Control enters attention: camera command.}
We use Pl\"ucker-ray conditioning~\cite{lfn}: each token is given the ray through
its pixel, expressed in the first frame's camera, as direction $\mathbf{d}$ and moment
$\mathbf{m}=\mathbf{o}\times\mathbf{d}$ (related encodings:~\cite{gta,prope,raynova}).
A small ray encoder maps them to offsets of the queries and keys,
$Q' = Q + \Delta Q(\mathbf{d},\mathbf{m})$ and $K' = K + \Delta K(\mathbf{d},\mathbf{m})$, of a
parallel self-attention branch in 8 transformer blocks, whose output starts at zero and is added
to the residual stream; the base model stays frozen (\cref{fig:arch}).
Offsets that condition only the control band give nearly all of the gain on natural camera paths (\cref{sec:exp:camera}).

\paragraph{Subject command: coarse-token transplant.}
The transplants of \cref{sec:method:control} also say how a subject can be commanded without
learning it: tokens copied into levels 0--3 carry their motion into the output, and the model
renders the rest. We therefore build a \emph{puppet} video of the command: cut the subject out of frame 0
along its mask, inpaint the hole~\cite{telea},
and then slide the cut-out rigidly, scaled along the commanded path similar to~\cite{ttm}.
The model's own tokenizer encodes the puppet, and during generation its tokens replace the
sampled ones at levels 0--3, over the whole frame. Exploiting the identified token hierarchy, 
we let the remaining lower levels generate the subject's appearance, hole and background,
without the model learning the subject's motion. 

Transplanting all the tokens of a frame from the motion control band 
copies the camera motion too. To spatially separate camera motion from the subject motion,
we introduce a region-transplant version instead of copying all the tokens from a level. 
The puppet is written only into the corresponding cells of the hierarchy (``region'' of the frame) 
that the subject occupies now or occupied at frame 0 (\cref{fig:redraw}). 
The control signal is trained with the same recipe 
as camera control (\cref{fig:arch}), with an added subject outline branch.

\paragraph{2. Memory is stored in the coarse levels.}
With one clip of context, a scene that leaves the view is forgotten. Informed by our hierarchy (\cref{sec:method:render}),
we keep the coarse levels (levels 0--10) excluding the detail band. 
This lets us hold the scene's committed content and motion, 
and lets the model regenerate the finest levels. 
The store is only 88\,KB, $11.6\times$ smaller than the full token pyramid.
On a return, we retrieve the stored clip whose cameras best match the requested path and force
its coarse levels into the new call.

The control map also says how \emph{not} to read the memory.
Because the camera is decided at levels 1--3, copying stored tokens as they are replays the
stored clip's motion: on a return path, the raw copy moves the camera the wrong way (\cref{sec:exp:memory}).
The reader therefore decodes the stored clip, re-orders its frames to match the new camera
path, re-encodes them in the current context and forces the resulting levels 0--10.
A plain decode of levels 0--10 is blurry, so we first \emph{redraw the detail} (\cref{fig:redraw}): one extra 5\,s
generation along the stored path, with levels 0--10 forced, regenerates levels 11--13 before the
frames are re-ordered.
The store stays 88\,KB; the reader has no learned parameters.

\section{Experiments}
\label{sec:exp}

We test the recipe with three different uses of the same map: camera control of
a static scene (\cref{sec:exp:camera}), control of a moving subject (\cref{sec:exp:subject}), and a
scene memory for videos longer than one clip (\cref{sec:exp:memory}).

\paragraph{Base model and training.}
Our model is built on InfinityStar~\cite{infinitystar}, to our knowledge the only publicly
available next-scale video model with discrete tokens.
We generate at $480$p, 5\,s per clip, keep its weights frozen and train with its bitwise loss
(\cref{eq:loss}); only the adapters and encoders of the recipe in \cref{fig:arch} are trained.
The camera adapter (\cref{sec:method:recipe}) is trained on RealEstate10K~\cite{re10k},
and the subject branch used by the region-select transplant
(\cref{sec:method:recipe}, \cref{fig:redraw}) on SA-V~\cite{sam2} clips. More details are in \cref{sec:supp_training,sec:supp_branches}.

\subsection{Camera control of a static scene}
\label{sec:exp:camera}

\paragraph{Test setup.}
We test on two types of camera motion data.
\emph{RealEstate10K:} the test split of the training dataset, with smaller peak deviations from the starting frame.
\emph{test100:} randomly sampled starting frames from the RealEstate10K test split, with 9 user-commanded
trajectories, including rotation-heavy (pan $90^\circ$ left/right, pan $45^\circ$, tilt $90^\circ$/$30^\circ$), 
zoom, and translation-heavy (pedestal, dolly, orbit) motions.
We re-measure the generated camera with MegaSAM~\cite{megasam} and report rotation error (RotErr, degrees) and
translation error (TransErr, normalized to RealEstate10K units). 

We compare against CameraCtrl~\cite{cameractrl} and Diffusion-as-Shader (DaS)~\cite{das},
which add camera control to continuous-latent video diffusion models.
We also include two recent methods outside this setting (see \cref{sec:supp_camera_details,sec:supp_other_discrete}):
AC3D~\cite{ac3d}, released as text-to-video, so only its camera metrics are comparable, and
RAYNOVA~\cite{raynova}, a next-scale model trained on driving video, which we fine-tune on our
training data until its camera error plateaus. All methods get the same path, resampled to their native frame counts.

\begin{table}[t]
  \centering\footnotesize
  \setlength{\tabcolsep}{1.4pt}
  \begin{tabular}{@{}llcccccc@{}}
    \toprule
    & & \multicolumn{2}{c}{\textbf{Ours}} & AC3D & CameraCtrl & DaS & RAYNOVA\\
    \cmidrule(lr){3-4}
    \multicolumn{2}{@{}r}{\emph{levels}} & all & 1--3 & \cite{ac3d} & \cite{cameractrl} & \cite{das} & \cite{raynova}\\
    \midrule
    \multicolumn{8}{@{}l}{\emph{test100} (100 clips)}\\
    Rotations    & RotErr\,$\downarrow$   & \textbf{11.10} & 13.51 & 15.10 & 22.46 & 30.17 & 38.00\\
    Zoom         & RotErr\,$\downarrow$   & \textbf{1.11}  & 2.25  & 2.26  & 1.74  & 5.81  & 15.55\\
    Transl.      & TransErr\,$\downarrow$ & \textbf{0.26} & 0.31 & 0.44 & 0.54 & 0.59 & 1.44\\
    \midrule
    \multicolumn{8}{@{}l}{\emph{RealEstate10K test split} (299 clips)}\\
    \multicolumn{2}{@{}l}{RotErr\,$\downarrow$} & \textbf{4.56} & 4.64 & 4.61 & 5.00 & 11.08 & 23.38\\
    \multicolumn{2}{@{}l}{TransErr\,$\downarrow$}      & \textbf{0.31} & 0.34 & 0.34 & 0.34 & 0.56 & 1.35\\
    \midrule
    \multicolumn{2}{@{}l}{Min.\ per 10\,s video\,$\downarrow$} & \multicolumn{2}{c}{\textbf{2.3}} & 7.9 & 3.9 & 12.1 & 8.2\\
    \multicolumn{2}{@{}l}{FVD (visual quality)\,$\downarrow$}                   & \textbf{591} & 620 & 1092 & 743 & 716 & 1108\\
    \bottomrule
  \end{tabular}
  \caption{\textbf{Camera control.} Errors of the generated camera,
  re-measured with MegaSAM~\cite{megasam} (RotErr in degrees, TransErr in RealEstate10K~\cite{re10k} units).
  Ours: camera adapter at all levels, or only at levels 1--3.
  Full table and videos in \cref{sec:supp_camera_details}.}
  \label{tab:camera_main}
\end{table}

\paragraph{Results.}
Our model with control at every level has the lowest camera error on both benchmarks, and is the fastest (\cref{tab:camera_main}).
Its advantage grows as the camera path departs from the RealEstate10K training distribution.
On the RealEstate10K test-split clips, we narrowly beat the next best baseline.
On the commanded test100 paths, however, the gap is larger: our rotation / translation errors are $26\%$ / $40\%$ 
lower than the next best, as illustrated in \cref{fig:camera}.
We posit that this stems from better camera control on out of distribution paths
from the models' training/finetuning, which indicates that our design generalizes better.
A per-motion breakdown of the commanded motions (\cref{tab:supp_camera_test100}) shows that the gap
is largest on the extreme pans. 

\begin{table}[t]
  \centering\footnotesize
  \setlength{\tabcolsep}{4pt}
  \begin{tabular}{@{}lcc@{}}
    \toprule
    Camera adapter active at & RotErr\,$\downarrow$ & TransErr\,$\downarrow$\\
    \midrule
    (A) levels 1--3 (control band, $0.7\%$ of tokens) & 6.95 & 0.270\\
    (B) every level               & \textbf{6.69} & \textbf{0.235}\\
    (C) levels 5--13 ($98.5\%$ of tokens) & 18.41 & 0.812\\
    No camera (adapter off)       & 18.02 & 0.725\\
    \bottomrule
  \end{tabular}
  \caption{\textbf{Placement ablation} of the camera adapters on
  40 held-out clips~\cite{re10k}. Control in the control band
  alone ($0.7\%$ of tokens) recovers nearly all of the every-level gain; at the levels where motion
  becomes visible, it is no better than no camera.}
  \label{tab:placement}
\end{table}

\paragraph{Where must camera control act?}
\label{sec:exp:placement}
We train camera adapters that differ only in where the conditiong acts (\cref{tab:placement}):
(A) the control band, levels 1--3; (B) every level; (C) levels 5--13, where motion becomes visible.
A recovers $98\%$ of B's rotation gain over no camera and $93\%$ of its translation gain, and the
two are not significantly different ($p{=}0.57$).
Control can thus act on $0.7\%$ of the tokens for nearly full accuracy on natural camera paths, or at every
level for the most accurate large commanded motions (\cref{tab:camera_main}).

C is no better than no camera: it generates the same levels 0--4 whatever the command,
and the finer levels cannot impose a camera path on them, although the command shifts their
predictions as much as in the full model. This supports the principle behind our recipe in \cref{sec:method:recipe}: 
\emph{Control signal must act before motion appears} at level 5. More details are in \cref{sec:supp_placement}.

\subsection{Subject control}
\label{sec:exp:subject}

\paragraph{Setup.}
We command a subject with the region-token transplant of \cref{sec:method:recipe} on our camera
model, on 14 DAVIS~\cite{davis2017} sequences with one clearly moving subject.
Each first frame gets the subject's own path, the
same path with a pause (move--stop--move), and stay. For a moving camera, the camera also turns
towards where the subject walks, by half or all of its sideways travel. We track the subject with
SAM\,2~\cite{sam2} and remove camera motion with background homographies.
\emph{Err.} is ObjMC~\cite{motionctrl} with camera motion removed and divided by the path length;
\emph{Pause} is the motion during the pause. \emph{Restart} counts the sequences in
which the subject moves more after the pause than during it. \emph{Held} counts stay commanded paths that remain 
within $2\%$ of the frame width.
We compare with Time-to-Move (TTM)~\cite{ttm},
which also steers a puppet on Wan 2.2~\cite{wan} and
SVD-xt~\cite{svd}, and with SG-I2V~\cite{sgi2v}; all get our puppet, path and caption.

\paragraph{Results.}
With a still camera (\cref{tab:subject_still}), ours matches TTM on the subject's own
path, equal to the ceiling that transplants the real clip's levels 0--3,
and performs better on move-stop-move commands, all at $15\times$ the speed.

The harder test is when the camera turns with the moving subject (\cref{tab:subject_pan}). 
Although both follow the camera, our subject error stays near its still-camera level while TTM's
rises (worse than still in 8/28 runs vs.\ 20/28). 
Our model thus follows a camera and a subject command at the same time, keeping the subject
closer to its path than TTM in all paired comparisons.
\Cref{sec:supp_subject_pan} lists these comparisons and shows that both choices of \cref{sec:method:recipe}
contribute to the gain: without the subject branch, the subject error under the turn rises to TTM's level,
and without the region transplant, the camera misses the commanded turn. Writing the transplant only into
the control band also avoids pasting the puppet and mitigates inpainted holes.

\begin{table}[t]
  \centering
  \setlength{\tabcolsep}{3pt}
  \resizebox{\linewidth}{!}{%
  \begin{tabular}{@{}lcccccc@{}}
    \toprule
    & Own path & \multicolumn{3}{c}{Move--stop--move} & Stay & Time\\
    \cmidrule(lr){2-2}\cmidrule(lr){3-5}\cmidrule(lr){6-6}\cmidrule(l){7-7}
    Method & Err.\,$\downarrow$ & Err.\,$\downarrow$ & Pause\,$\downarrow$ & Restart\,$\uparrow$ & Held\,$\uparrow$ & (s)\,$\downarrow$\\
    \midrule
    TTM~\cite{ttm}, Wan 2.2 A14B & \textbf{0.05} & 0.08 & 0.09 & 8/14 & \textbf{11/14} & 931\\
    SG-I2V~\cite{sgi2v}, SVD     & 0.36 & 0.27 & \textbf{0.02}$^\dagger$ & 7/14 & 7/14 & 142\\
    Ours: whole-frame transplant (no training) & 0.07 & 0.07 & 0.05 & 11/14 & \textbf{11/14} & 60\\ %
    \textbf{Ours: region transplant + subject branch} & 0.06 & \textbf{0.04} & \textbf{0.04} & \textbf{12/14} & \textbf{11/14} & 60\\
    \midrule
    \rowcolor{refgrey} Ceiling: real clip's levels 0--3 & 0.06 & -- & -- & -- & -- & 60\\
    \bottomrule
  \end{tabular}}
  \caption{\textbf{Subject control, still camera,} 14 DAVIS~\cite{davis2017} sequences (medians).
  Time per 5\,s video on one A100. $^\dagger$SG-I2V barely moves the subject. Videos on the \projectpage.}
  \label{tab:subject_still}
\end{table}
\begin{table}[t]
  \centering\footnotesize
  \setlength{\tabcolsep}{1.4pt}
  \begin{tabular}{@{}lcccc@{}}
    \toprule
    & \multicolumn{2}{c}{Err.\,$\downarrow$} & Worse than & Camera\\
    \cmidrule(lr){2-3}
    Method & half & full & still\,$\downarrow$ & reach (full)\\
    \midrule
    TTM~\cite{ttm}, camera recipe & 0.10 & 0.12 & 20/28 & 0.97\\ %
    TTM~\cite{ttm}, cut-and-drag  & 0.09 & 0.10 & 20/28 & 0.93\\ %
    TTM, SVD-xt, camera mask       & 0.36 & 0.48 & 20/28 & 0.03\\ %
    TTM, SVD-xt, cut-and-drag      & 0.40 & 0.47 & 21/28 & 0.01\\ %
    SG-I2V~\cite{sgi2v}, + backgr.\ boxes & 0.32 & 0.47 & 17/28 & 0.20\\ %
    SG-I2V~\cite{sgi2v}, subject box & 0.50 & 0.51 & 25/28 & 0.08\\ %
    \textbf{Ours: reg. transplant + subj. branch}   & \textbf{0.04} & \textbf{0.07} & \textbf{8/28} & 0.95\\ %
    \bottomrule
  \end{tabular}
  \caption{\textbf{Subject control with a moving camera}, own path, same 14 sequences. The camera
  turns towards where the subject walks, by half or all of its sideways travel. Worse than still:
  runs whose error exceeds the same method's still-camera error. Camera reach: measured / commanded turn, on the full turn.}
  \label{tab:subject_pan}
\end{table}

\subsection{Memory for longer videos}
\label{sec:exp:memory}

Our hierarchy tells us what to store and how to read it. We test each choice, then compare
the resulting store with models trained for memory.

\begin{figure*}[t]
  \centering
  \includegraphics[width=\linewidth]{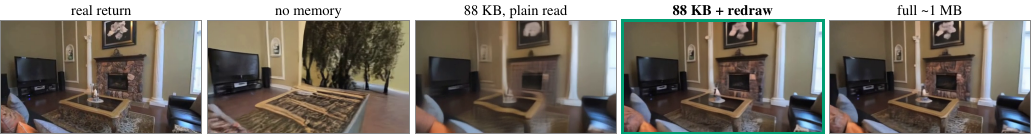}%
  \caption{\textbf{Turning back in a living room.} Held-out RealEstate10K~\cite{re10k} return. Without memory, a new room appears;
  the 88\,KB store restores it, blurry; redrawing the detail sharpens it (25.0 vs.\ 22.5\,dB).}
  \label{fig:memory}
\end{figure*}

\begin{figure}[t]
  \centering
  \includegraphics[width=\linewidth]{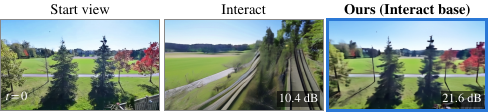}%
  \caption{\textbf{Coming back after 20\,s.} After turning right, back, left and back, the frame at $t{=}20$\,s
  should match the start view (dB: PSNR to it). Interact alone does not come back. With our adapter,
  store and redraw, it does. Held-out RealEstate10K~\cite{re10k} scene.}
  \label{fig:mem_baselines}
\end{figure}

\paragraph{Setup.}
Each test stores an outward clip and scores the return to it.
\emph{Real returns} (5\,s; \cref{tab:mem_phase1,fig:memory}): 8 held-out RealEstate10K~\cite{re10k}
clips, four rotation- (Rot.) and four translation-dominant (Trans.), split into an outward and a return
leg and scored by PSNR against the real return footage.
\emph{Generated chains} (20\,s; \cref{tab:mem_phase2,tab:mem_baselines,fig:mem_baselines}): 8 held-out
RealEstate10K scenes, four 5\,s calls (pan right, back, left, back). Each return is scored against the
model's own outward frames at the same pose by PSNR, LPIPS and camera error, the yaw left after a turn and
its return (from RAFT~\cite{raft} flow); Turns counts the scenes whose four legs all turn the commanded way.
\Cref{tab:mem_phase1} also scores three harder generated revisits (8 scenes each): a return that
overlaps the stored clip only partly, one that spans two stored clips, and one twice as fast.
We compare with UCM~\cite{ucm}, Spatia~\cite{spatia} and
InfinityStar-Interact~\cite{infinitystar}.

\begin{table}[t]
  \centering\small
  \setlength{\tabcolsep}{2.5pt}
  \resizebox{\linewidth}{!}{%
  \begin{tabular}{@{}lcccccc@{}}
    \toprule
    & & \multicolumn{2}{c}{Real returns} & \multicolumn{3}{c}{Generated revisits}\\
    \cmidrule(lr){3-4}\cmidrule(l){5-7}
    Store of the outward clip & Size & Rot. & Trans. & Partial & Two clips & 2$\times$ fast\\
    \midrule
    \rowcolor{refgrey} No memory          & 0       & 11.6 & 12.0 & 13.1 & 12.9 & 12.1\\
    Levels 0--10, raw copying               & 88\,KB  & 11.9 & 12.5 & 12.1 & 11.6 & 11.5\\
    \textbf{Levels 0--10 (ours)}  & \textbf{88\,KB} & \textbf{21.8} & \textbf{21.9} & \textbf{21.9} & \textbf{22.6} & \textbf{21.1}\\
    All levels                           & ${\sim}1$\,MB & 25.4 & 24.7 & 24.9 & 26.3 & 23.3\\
    \rowcolor{refgrey} Oracle             & --      & 25.5 & 24.9 & -- & -- & --\\
    \bottomrule
  \end{tabular}}
  \caption{\textbf{Coarse levels restore a revisited view.} PSNR (dB) of the return leg against
  real RealEstate10K~\cite{re10k} videos. Raw token copying fails.
  Our plain read (reorder+ reencode, without redrawing detail) recovers most of the gain of an $11.6\times$ larger store.
  Generated revisits do not retrace the stored clip and are scored against the model's own frames.}
  \label{tab:mem_phase1}
\end{table}

\begin{table}[t]
  \centering\small
  \setlength{\tabcolsep}{3.5pt}
  \begin{tabular}{@{}lccc@{}}
    \toprule
    Store (20\,s chain, 8 scenes) & Size & Gain (dB) $\uparrow$ & Return err.\ $\downarrow$\\
    \midrule
    \textbf{Discrete levels 0--10} & 88\,KB & +10.2 & 5.2 / 3.1$^\circ$\\
    \quad\textbf{+ redraw detail}  & 88\,KB & \textbf{+10.3} & \textbf{0.6 / 0.7$^\circ$}\\
    JPEG frames                   & 84\,KB & +8.6 & 17 / 15$^\circ$\\
    Continuous latent              & 75\,KB & +6.9 & 30 / 17$^\circ$\\
    \midrule
    \rowcolor{refgrey} Lossless frames & 99\,MB & +11.0 & --\\
    \bottomrule
  \end{tabular}
  \caption{\textbf{At equal size, coarse tokens gain more than image or latent stores.}
Gain in return PSNR over no memory of the 20\,s chain.
Our 88\,KB store is above JPEG frames of the same size,
and reaches $93\%$ of the gain of lossless frames.}
  \label{tab:mem_phase2}
\end{table}

\begin{table}[t]
  \centering\scriptsize
  \setlength{\tabcolsep}{1.7pt}
  \begin{tabular}{@{}lccccccc@{}}
    \toprule
    & \multicolumn{4}{c}{Return to the stored view} & Cost & \multicolumn{2}{c}{Video quality}\\
    \cmidrule(lr){2-5}\cmidrule(lr){6-6}\cmidrule(l){7-8}
    Method & PSNR$\uparrow$ & LPIPS$\downarrow$ & Err.$\downarrow$ & Turns & min$\downarrow$ & FVD$\downarrow$ & Aes.$\uparrow$\\
    \midrule
    \rowcolor{refgrey} Ours, no memory & 12.9/12.8 & 0.510 & 7.5/11.7 & 8/8 & 3.8 & 982 & 42.9\\
    Interact~\cite{infinitystar}$^\dagger$ & 10.7/13.2 & 0.631 & 50/27 & 0/8 & 4.1 & 1072 & 44.7\\
    Spatia~\cite{spatia}$^\dagger$ & 18.4/17.9 & 0.262 & 1.6/2.2 & 7/8 & 47.5 & \textbf{778} & 44.2\\
    UCM~\cite{ucm}$^\dagger$ & 21.8/22.9 & 0.192 & \textbf{0.3}/1.0 & 8/8 & 14.1 & 850 & 44.8\\
    \midrule
    \textbf{Ours} (plain) & 22.8/23.6 & 0.151 & 0.6/0.7 & 8/8 & 5.9 & 958 & 48.0\\
    \textbf{Ours} + Interact base & \textbf{26.8/27.9} & \textbf{0.119} & 0.8/\textbf{0.5} & 8/8 & 6.2 & 991 & \textbf{49.4}\\
    \bottomrule
  \end{tabular}
  \caption{\textbf{A training-free 88\,KB memory returns as accurately as models trained for memory ($^\dagger$).}
  Ours: 88\,KB store + redraw. Per return (first / second): PSNR, camera error ($^\circ$); LPIPS averaged over both; 
  Aes: VBench aesthetic quality.}
  \label{tab:mem_baselines}
\end{table}

\paragraph{What to store.}
The layout band of our hierarchy ends at level 10 (\cref{sec:method:render}), while the
detail band holds $82\%$ of the tokens (\cref{sec:prelim}), so we store levels 0--10 and let the
model redraw the rest.
On real returns (\cref{tab:mem_phase1}) this gains about $10$\,dB over no memory in every motion
class, most of the gain of storing all levels at $11.6\times$ the size; keeping fewer levels loses
in every revisit we tested, and at equal size the store gains more than JPEG frames or a
continuous latent.

\paragraph{How to read.}
Because the coarse levels also decide motion (\cref{sec:method:control}), copying the stored
tokens as they are replays the outward motion and barely helps (\cref{tab:mem_phase1}).
We therefore re-order them to the return path, and redraw the detail before re-encoding
(\cref{fig:redraw}): this improves every real return and closes $61\%$ of the PSNR gap and
$79\%$ of the LPIPS gap to the ${\sim}1$\,MB store at the same 88\,KB (\cref{fig:memory}).
The same reader handles returns that do not retrace the stored clip (partial overlap, a return
over two stored clips, a $2\times$ faster or tilted return, and a loop closed in the same
direction), gaining 11--14\,dB in every case (\cref{sec:supp_mem_revisits}).

\paragraph{Against models trained for memory.}
Our memory trains nothing, yet it returns to the stored view as accurately as models trained for memory 
(\cref{tab:mem_baselines}): on the base model its return PSNR is $0.8$\,dB above UCM
and $5$\,dB above Spatia at a similar camera error.
UCM and Spatia remember \emph{pixels} and place them in the new view through estimated depth,
so depth errors, holes and colour drift can enter each return.
Our store follows the hierarchy instead: layout is decided at the coarse levels, while the detail
band holds $82\%$ of the tokens (\cref{sec:prelim}), so we keep only the model's own coarse tokens
and force them at the levels that decide layout.
The same generator then redraws the same room, with no geometry to estimate.
Interact alone cannot be steered back. With our camera adapter, trained on DL3DV~\cite{dl3dv},
and the same store and redraw, it reaches the highest return PSNR in the comparison.
This shows that our store carries over to a second next-scale model (\cref{fig:mem_baselines}).

\section{Conclusion}
\label{sec:conclusion}
We asked where in a discrete coarse-to-fine video hierarchy motion is decided, and how
continuous control can use it.
Token transplants show that camera and subject motion are decided at
levels 1--3, several levels before motion becomes visible.
Control must therefore exploit the hierarchy: the signal must reach
these coarse levels, and control placed only where motion becomes visible does nothing.
The same map turns coarse tokens into a subject command and into a compact memory.
With one discrete model, we steer a camera with lower error than comparable diffusion models in
less time, keep a subject closer to its path than its diffusion counterpart while the camera
moves, at $15\times$ the speed, and restore revisited views from a training-free 88\,KB store
that returns as accurately as models trained for memory, with no 3D reconstruction.
The hierarchy that makes discrete autoregressive video fast is also what makes it controllable.

\clearpage
\twocolumn[\centering\Large\textbf{\thetitle}\\\vspace{0.5em}Supplementary Material\\\vspace{1.0em}]
\appendix
\addtocontents{toc}{\protect\setcounter{tocdepth}{-1}}
\section{Supplementary Overview}
\label{sec:supp_overview}
\addtocontents{toc}{\protect\setcounter{tocdepth}{2}}

\begingroup
\small
\makeatletter
\@starttoc{toc}
\makeatother
\endgroup

\textbf{Videos.} Video results for the camera, subject and memory experiments are on the project page (\projecturl), one tab each (``Camera control'', ``Subject control'', ``Return memory''). We encourage readers
to view them alongside the tables.

\section{Details on Related Work}
\label{sec:supp_related}

\paragraph{Long, interactive video models.}
A world model must keep producing video while the user acts, so it cannot be a fixed-length
generator.
Most long-video systems therefore generate clip by clip, conditioning each clip on what was
already shown and caching keys and values so that the next clip streams~\cite{selfforcing,
matrixgame2}.
Interactive world models built this way take keyboard actions~\cite{matrixgame2}, latent
actions learned from video~\cite{genie} or camera commands~\cite{cosmos}, and are evaluated on
how quickly and consistently they respond.
Three requirements follow, and they frame this paper: the generator must have \emph{low latency}, be
\emph{continuously controllable} by more than a language prompt, and keep \emph{memory} of the scene for longer than one clip, ideally without
building an explicit 3D world that would slow it down.

\paragraph{Hierarchies and where generators decide things.}
In GANs, the observation that early layers set pose and later layers set texture made style
mixing and layer-wise control possible~\cite{stylegan}; in image animation, restricting motion
to a low-frequency band shaped the generative model~\cite{gid}.
In next-scale image models, coarse levels have been reused to preserve layout during
editing~\cite{aredit,dni}; in next-scale video models, fine-level tokens have been pruned
for speed~\cite{faststar}.
Unlike these, we measure where a property is \emph{decided}, separate it from when it becomes
visible, and show that the two differ.

\section{Hierarchy Is the Key: Details}
\label{sec:supp_method}

\subsection{Next-scale prediction and clip structure}
\label{sec:supp_nextscale}
\begin{figure}[t]
  \centering
  \includegraphics[width=\linewidth]{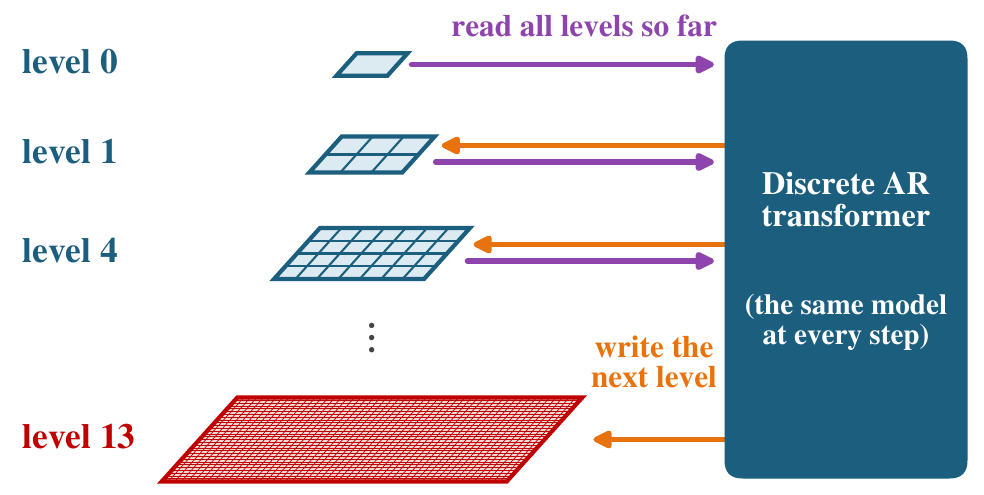}
  \caption{\textbf{Next-scale prediction writes a clip coarse to fine.}
  The transformer writes the clip level by level; each step reads every coarser level of the clip and
  writes the next one. Each level refines the one above it: coarse levels set the scene, fine
  levels add detail.}
  \label{fig:nextscale}
\end{figure}

\begin{figure}[t]
  \centering
  \includegraphics[width=\linewidth]{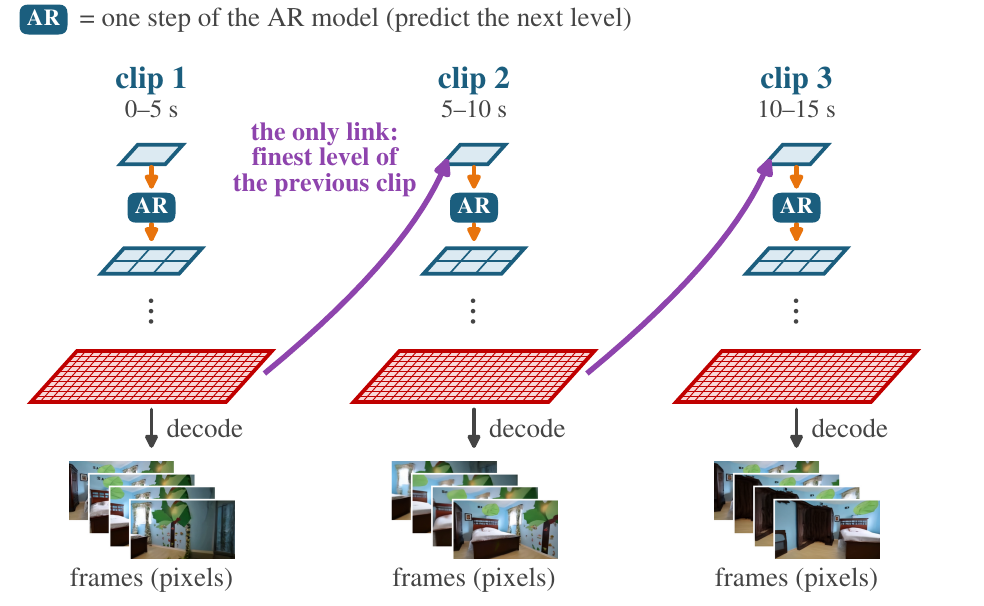}
  \caption{\textbf{Clip structure.} Video is made in 5\,s clips; every level of clip $n$
  attends only to the finest level of clip $n{-}1$. Older clips are not attended to directly,
  so a view that stays out of frame for longer may be forgotten.}
  \label{fig:clips}
\end{figure}

This section gives the detailed versions of the next-scale overview figure in the main text
(\cref{fig:nextscale_combined}):
how one clip is generated level by level (\cref{fig:nextscale}), the token grid and token count
of every level (\cref{tab:levels}), and how clips are chained (\cref{fig:clips}).

\begin{table}[t]
  \centering\small
  \setlength{\tabcolsep}{3.2pt}
  \begin{tabular}{@{}lccccccc@{}}
    \toprule
    Level & 0 & 1 & 2 & 3 & 4 & 5 & 6\\
    Grid  & $1{\times}1$ & $2{\times}3$ & $2{\times}4$ & $3{\times}5$ & $4{\times}7$ & $4{\times}8$ & $5{\times}9$\\
    Tokens & 20 & 120 & 160 & 300 & 560 & 640 & 900\\
    \midrule
    Level & 7 & 8 & 9 & 10 & 11 & 12 & 13\\
    Grid  & $6{\times}11$ & $8{\times}13$ & $9{\times}16$ & $12{\times}21$ & $18{\times}32$ & $24{\times}43$ & $30{\times}53$\\
    Tokens & 1320 & 2080 & 2880 & 5040 & 11520 & 20640 & 31800\\
    \bottomrule
  \end{tabular}
  \caption{\textbf{Levels 1--3, where we will find the camera is decided, hold only $0.7\%$ of a
  clip's tokens; levels 11--13 hold $82\%$.} Token grid per latent frame and tokens per
  5\,s clip ($T{=}20$).}
  \label{tab:levels}
\end{table}

\paragraph{Token representation and residual quantization.}
An autoencoder maps each 5\,s clip to a latent
$z \in \mathbb{R}^{T\times H\times W\times d}$ with $T{=}20$ latent frames.
A multi-level residual quantizer~\cite{var,bsq,infinity,infinitystar} writes it as $K{=}14$ token maps
$r_0,\dots,r_{13}$.
Level $k$ has a grid of $h_k\times w_k$ tokens per latent frame, from $1\times1$ at level 0
to $30\times53$ at level 13. Each token is a vector of binary codes: 16 bits at levels 0--10
and 64 bits at levels 11--13~\cite{bsq,infinity,infinitystar}.
Each level quantizes the residual left unexplained by coarser levels,
\begin{equation}
  r_k = \mathcal{Q}\Big(\mathrm{down}_k\big(z - \textstyle\sum_{j<k}\mathrm{up}(r_j)\big)\Big),
  \label{eq:supp_residual}
\end{equation}
so the decoder receives $\sum_k \mathrm{up}(r_k)$ and each level refines, but does not revise,
the earlier levels.
The full pyramid contains 77{,}980 tokens per clip. Levels 0--4 hold $1.5\%$ of the tokens,
while levels 11--13 hold $82\%$ (\cref{tab:levels}).

\paragraph{Next-scale factorization and training weights.}
The transformer predicts the maps in order,
\begin{equation}
  p(r_0,\dots,r_{13}\mid c) = \prod_{k=0}^{13} p\big(r_k \mid r_{<k},\, c\big),
  \label{eq:supp_nextscale}
\end{equation}
where $c$ contains text and the first frame, and for later clips also the finest level of the
previous clip. All tokens at a level are predicted in parallel, and each bit is predicted independently
given the context~\cite{infinity}; thus a clip takes a few dozen forward passes rather
than one per token.
In the BCE objective of the main text, \cref{eq:loss}, the per-token level weight is
$\lambda_k=\min\big(40,\sqrt{n_{13}/n_k}\big)$, where $n_k$ is the number of tokens at level
$k$. The total weight at a level therefore grows as $\sqrt{n_k}$, so the finest levels still
dominate the loss.

\subsection{Render bands}
\label{sec:supp_bands}
\begin{figure*}[t]
  \centering
  \includegraphics[width=\linewidth]{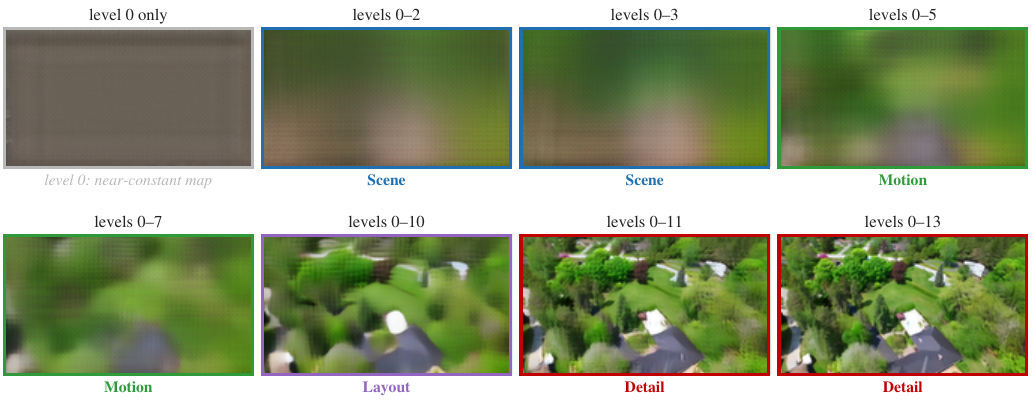}
  \caption{\textbf{What each level adds.} One held-out RealEstate10K~\cite{re10k} clip (aerial), frame 60, decoded from
  levels $0..k$ only. Below level ${\sim}4$ the decodes are blurred and carry decoder
  artifacts; by level 10 the layout is complete, and levels 11--13 add texture and
  detail. Visibility, however, is not where things are decided (\cref{sec:method}).}
  \label{fig:strip}
\end{figure*}

\begin{table}[t]
  \centering\small
  \setlength{\tabcolsep}{3pt}
  \begin{tabular}{@{}l p{0.75\linewidth}@{}}
    \toprule
    Band & Property and measure\\
    \midrule
    Scene  & \emph{Latent identity}: the cumulative latent after level $k$, average-pooled to
             $5{\times}6{\times}10$, must pick out its own clip among the 80; cosine similarity to
             its own clip's latent minus the highest similarity to any other clip.\\
    Motion & \emph{Flow direction}: RAFT-small~\cite{raft} optical flow average-pooled to a
             $6{\times}10$ grid; cosine similarity of the flattened flow fields to those of the
             reference clip.\\
    Layout & \emph{Low-resolution LPIPS}: LPIPS~\cite{lpips} (AlexNet) after area downsampling to
             $120{\times}212$.\\
    Detail & \emph{High-frequency energy}: mean squared response of a $3{\times}3$ Laplacian on
             the grayscale frame, as a ratio to that of the reference clip.\\
    \bottomrule
  \end{tabular}
  \caption{\textbf{Each render band is measured by one property of the partial decode.} Each measure is
  computed on every partial pyramid $0..k$ and averaged over the 80 clips.}
  \label{tab:supp_render_measures}
\end{table}

\begin{table*}[t]
  \centering\small
  \setlength{\tabcolsep}{3.4pt}
  \begin{tabular}{@{}l *{14}{c}@{}}
    \toprule
    Level $k$ & 0 & 1 & 2 & 3 & 4 & 5 & 6 & 7 & 8 & 9 & 10 & 11 & 12 & 13\\
    \midrule
    Latent identity (Scene)   & 0.00 & 0.11 & 0.34 & 0.46 & \textbf{0.62} & 0.80 & 0.86 & 0.92 & 0.97 & 0.97 & 0.98 & 1.00 & 1.00 & 1.00\\
    Flow direction (Motion)   & 0.00 & 0.03 & 0.03 & 0.20 & 0.33 & \textbf{0.51} & 0.63 & 0.74 & 0.87 & 0.92 & 0.95 & 0.99 & 1.00 & 1.00\\
    Low-res.\ LPIPS (Layout)  & 0.00 & 0.02 & 0.05 & 0.10 & 0.14 & 0.20 & 0.30 & 0.38 & \textbf{0.53} & 0.61 & 0.71 & 0.94 & 0.97 & 1.00\\
    High-freq.\ energy (Detail) & {\color{gray}0.96} & {\color{gray}0.48} & {\color{gray}0.13} & {\color{gray}0.04} & 0.00 & 0.08 & 0.10 & 0.14 & 0.19 & 0.23 & 0.30 & \textbf{0.72} & 0.86 & 1.00\\
    \bottomrule
  \end{tabular}
  \caption{\textbf{Motion becomes visible only at level 5, after scene identity (level 4) and before
  layout (8) and detail (11).} Render-band progress per level (0 = worst value, 1 = the full decode), mean
  over 80 held-out RealEstate10K clips. Bold: the first level past $0.5$, the onset of the property's
  band. High-frequency energy is counted from its minimum at level 4; gray: out-of-distribution
  decodes below it.}
  \label{tab:supp_render}
\end{table*}

This section details the render bands of the main text (\cref{sec:method:render}): the clips, how
each property is measured, and where each band begins.

\paragraph{Partial decoding.}
Decoding only levels $0..k$ of a held-out clip shows what each level adds
(\cref{fig:strip}). Below about level 4, the decodes are blurred and carry decoder artifacts.
The layout is in place by level 10, while levels 11--13 add texture and detail.

\paragraph{Clips.}
We use 80 held-out RealEstate10K~\cite{re10k} clips: 40 with rotation-dominant camera motion, 20
with strong translation and 20 mixed. Each is one 5\,s clip (81 frames), tokenized with the
posterior mean and decoded from levels $0..k$ for every $k$. Frame measures use every fourth
frame (20 frames); flow uses 10 frame pairs, four frames apart.

\paragraph{Properties.}
\Cref{tab:supp_render_measures} lists how each property is measured. Scene identity is measured
on the partial latent rather than its decode, because decodes below level 4 are blurred.

\paragraph{Progress and onsets.}
Each property's score at level $k$ is compared with its score for the full decode (levels
$0..13$) through its \emph{progress}, the fraction of its total change that levels $0..k$
cover:
$p_k = (s_k - s_{\min}) / (s_{13} - s_{\min})$, with $s$ oriented so that higher is better
(LPIPS is negated) and $s_{\min}$ its worst value over the levels
(\cref{tab:supp_render}). Measuring from the worst value only changes high-frequency energy, whose
worst value is at level 4: the decodes below it carry decoder artifacts.
A property's onset is the level where it first passes $50\%$ of its final value: latent
scene identity passes at level 4, flow direction at 5, low-resolution LPIPS at 8 and
high-frequency energy at 11.
Flow direction, low-resolution LPIPS and high-frequency energy start the Motion (5--7), Layout
(8--10) and Detail (11--13) bands. Scene is the first band, from level 1 up to its onset at level 4;
level 0 is a near-constant 20-token map.
Informed by this hierarchy, the memory keeps the coarse levels 0--10 and leaves the Detail band to be
regenerated (main text, \cref{sec:method:recipe}).
This describes where properties become visible; it does not by itself show where they
are decided (\cref{sec:supp_control}).

\subsection{Control bands}
\label{sec:supp_control}
\begin{figure}[t]
  \centering
  \includegraphics[width=\linewidth]{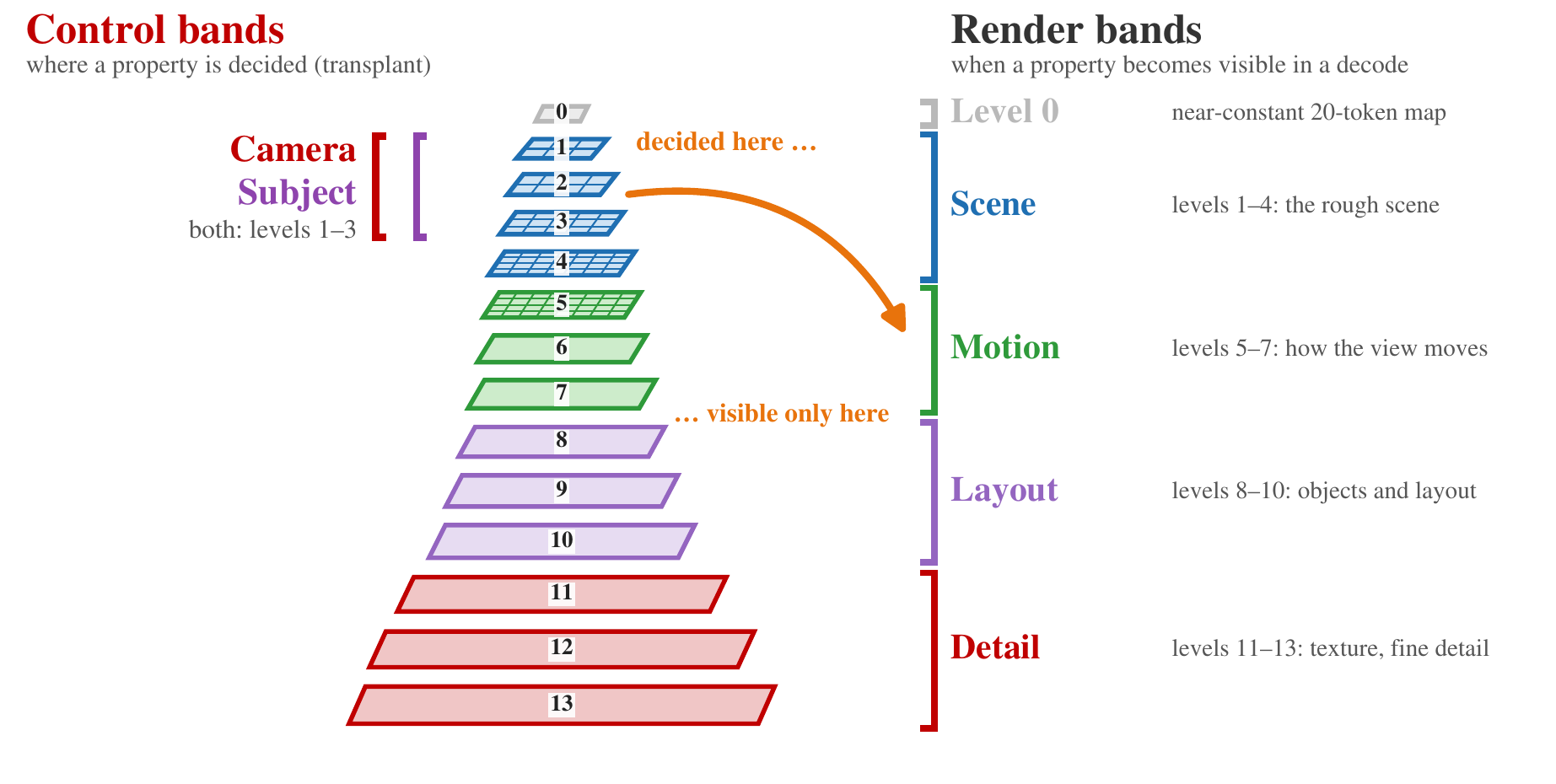}
  \caption{\textbf{Render bands and control bands do not coincide.}
  Right: render bands, measured on the tokenizer, mark when each property becomes visible in a
  partial decode (Scene 1--4, Motion 5--7, Layout 8--10, Detail 11--13; level 0 is a near-constant
  20-token map).
  Left: control bands, measured by token transplant, mark where it is decided.
  Camera and subject motion are both decided at levels 1--3,
  before the Motion band, where motion first becomes visible (levels 5--7).}
  \label{fig:bands}
\end{figure}

This section details the control bands of the main text (\cref{sec:method:control}, \cref{fig:transfer}).
\Cref{fig:bands} places them next to the render bands of \cref{sec:supp_bands}.

\paragraph{Control-band transfer results.}
Table~\ref{tab:supp_transplant} reports the camera and subject transfer values used to locate
their control bands. Transfer is measured by copying levels $0..m$ from a donor generation into
a recipient (0 = recipient, 1 = donor).

\begin{table}[t]
  \centering\small
  \setlength{\tabcolsep}{3.2pt}
  \begin{tabular}{@{}lcccccc@{}}
    \toprule
    Levels copied $0..m$ & 1 & 2 & 3 & 5 & 7 & 10\\
    \midrule
    \multicolumn{7}{@{}l}{\textit{Camera, standard commands (20 scenes)}}\\
    pan $45^\circ\rightarrow$ stay still   & 0.70 & \textbf{0.97} & 0.97 & 1.00 & 1.00 & 1.00\\
    pan $45^\circ\rightarrow$ opposite pan & 0.54 & \textbf{0.97} & 0.99 & 1.00 & 1.00 & 1.00\\
    tilt $30^\circ\rightarrow$ stay still  & 0.70 & \textbf{0.91} & 0.96 & 1.00 & 1.00 & 1.00\\
    zoom $\rightarrow$ stay still          & 0.82 & \textbf{0.96} & 0.99 & 1.00 & 1.00 & 1.00\\
    dolly $\rightarrow$ stay still         & 0.77 & \textbf{0.93} & 0.94 & 0.98 & 0.99 & 1.00\\
    PSNR to donor (dB)                     & 14.7 & 18.0 & 20.4 & 23.5 & 26.2 & 32.0\\
    \midrule
    \multicolumn{7}{@{}l}{\textit{Camera, large commands (20 scenes)}}\\
    pan $90^\circ\rightarrow$ stay still  & 0.46 & 0.93 & \textbf{0.98} & 0.99 & 1.00 & 1.00\\
    tilt $90^\circ\rightarrow$ stay still & 0.34 & 0.90 & \textbf{0.96} & 0.98 & 1.00 & 1.00\\
    pedestal $\rightarrow$ stay still      & 0.60 & 0.85 & \textbf{0.95} & 0.99 & 0.99 & 1.00\\
    \midrule
    Positions copied                       & 140 & 300 & 600 & 1.8k & 4.0k & 14.0k\\
    \midrule
    \multicolumn{7}{@{}l}{\textit{Subject, still camera (66 DAVIS sequences)}}\\
    walk $\rightarrow$ stay still          & 0.45 & 0.83 & 0.87 & 0.97 & 0.99 & 1.00\\
    stay still $\rightarrow$ walk          & 0.55 & 0.86 & 0.95 & 1.00 & 0.99 & 1.00\\
    \bottomrule
  \end{tabular}
  \caption{\textbf{Three coarse levels decide camera and subject motion: copying levels 0--2 (300 of
  77{,}980 positions) transfers a donor's standard camera motion, and levels 0--3 its largest.}
  Transfer of motion from a donor generation into a recipient when levels $0..m$ are copied.
  Standard motions transfer even against an opposite command, while the output still differs from the
  donor in content (PSNR, standard commands).
  Positions: token positions copied, of 77{,}980 per 5\,s clip. Camera rows: mean over the 20 scenes
  of each command. Subject rows show transfer of the subject's final position (median over
  sequences); most motion is decided at $m{=}2$, and 54 of the 66 sequences are past half at $m{=}3$.
  Main-text \cref{fig:transfer} pools the camera rows over all scene--command clips and plots
  their median with a 95\% bootstrap confidence interval.}
  \label{tab:supp_transplant}
\end{table}

\paragraph{Camera motion.}
We test our camera model (\cref{sec:method:recipe,sec:exp:camera}) on 20 held-out
RealEstate10K scenes, using standard benchmark commands and larger motions.
Donor and recipient share the first frame, caption and seed; their camera commands are given to
our camera adapter as rays.
Copying levels 0--1, or 140 token positions, into a call told to stay still carries
$0.70$--$0.82$ of a standard motion.
$0.91$--$0.97$ of standard pans, tilts, zooms and dollies (median over scene--command clips $0.96$;
96\% of clips at $0.8$ or more).
The large commands need level 3 as well: from levels 0--2 to 0--3, transfer rises from $0.93$ to
$0.98$ for a $90^\circ$ pan, from $0.90$ to $0.96$ for a $90^\circ$ tilt, and from $0.85$ to $0.95$
for a pedestal; levels 0--3 contain 600 positions, and the band, levels 1--3, holds $0.7\%$ of the
clip's 77{,}980 tokens. The decision also holds against an opposite command: a call told to pan left,
given the coarse tokens of a right pan, pans right in all 20 scenes (flow-field cosine $0.98$ to
the donor and $-0.94$ to the commanded motion). Motion transfers before content does: at $m{=}2$
the output moves like the donor but matches its pixels at only $18.0$\,dB, compared with
$32.0$\,dB at $m{=}10$.

\paragraph{Subject motion.}
We repeat the transplant with our subject branch (\cref{sec:exp:subject})
and a still camera on all 80 pre-registered DAVIS~\cite{davis2017} movers, which include the 14 sequences of the main subject experiments. The donor follows the subject's real path, while the
recipient stays in place; SAM\,2~\cite{sam2} tracks the subject from the human-annotated frame-0
mask, with residual camera motion compensated. On the 66 sequences where donor and recipient
end at least $0.02$ frame widths apart (the others barely move), copying levels 0--2 carries $0.83$ of the walk into the
recipient, and levels 0--3 carry $0.87$; 45 sequences are past half at $m{=}2$ and 54 at
$m{=}3$ (\cref{tab:supp_transplant}). In the reverse test, a walking call given the stay's
levels 0--3 stops ($0.95$). Most subject motion is therefore decided by levels 1--2, with level
3 settling it; transfer reaches $0.97$ by $m{=}5$.
The subject control band coincides with the camera's, so the two must be separated spatially,
not by level.

\begin{table}[t]
  \centering\small
  \setlength{\tabcolsep}{2.6pt}
  \begin{tabular}{@{}lccccc@{}}
    \toprule
    Levels forced & 0 & 1 & 1--2 & 1--3 & 3\\
    \midrule
    Camera, our model                  & 0.00 & 0.73 & \textbf{0.98} & 1.01 & 0.13\\
    \quad large commands               & 0.01 & 0.40 & \textbf{0.94} & 0.95 & 0.08\\
    Camera, base model$^\ast$          & --   & 0.99 & \textbf{1.00} & 1.00 & --  \\
    Camera, arm B$^\ast$               & --   & 0.86 & \textbf{0.99} & 0.98 & --  \\
    Subject, 66 DAVIS movers$^\ast$    & 0.00 & 0.45 & \textbf{0.83} & 0.87 & --  \\
    \bottomrule
  \end{tabular}
  \caption{\textbf{Levels 1--3 together carry the motion, also in the base model without an adapter;
  level 3 alone does not.} Motion transfer (0 = recipient, 1 = donor) when only
  the listed levels are forced. $^\ast$Level 0 is forced as well (it carries no motion, first column).
  Base model: the adapter switched off at every level. Arm B: adapter at every level
  (\cref{sec:supp_placement}). Medians; values above 1 overshoot the donor slightly.}
  \label{tab:supp_bandchecks}
\end{table}

\paragraph{Further checks of the control band.}
Three more inference-only tests support the band (\cref{tab:supp_bandchecks}; a fourth, switching the
adapter off outside chosen levels, is in \cref{tab:supp_gating}).
Forcing only chosen levels of our model, level 0 alone carries nothing and level 3 alone
little ($0.13$), while levels 1--2 carry as much as levels 0--2 do.
With the adapter switched off at every level, the frozen base model follows the donor's coarse
tokens completely ($1.00$ at levels 0--2), so the band is not an artifact of the adapter: the
decision is made in the backbone's own coarse levels and the adapter's role is to write it there.
Arm B, trained with the adapter at every level, transfers as our model does ($0.99$).
The subject row repeats the 66 movers of \cref{tab:supp_transplant} for comparison.

\paragraph{Logit probe: where the command acts.}
Transplants show where motion is decided; a logit probe shows where the command changes the
model's predictions. On 8 held-out RealEstate10K scenes, we force every token of a run to match
the ``stay still'' run, switch the command, and measure the change $|\Delta p|$ in each bit's
predicted probability at each level.
The command changes probabilities at every level except 0, but only at levels 1--2 does the
effect grow over the clip like a camera path (for a $45^\circ$ pan, $7\times$ from the first to
the last latent frame). At levels 3--7 the effect is nearly flat, and at levels 8--13 it is a
near-constant re-rendering.
The probe thus adds evidence for levels 1--2, and the flat effect at level 3 is consistent with
level 3 being needed only for the largest commands (\cref{tab:supp_transplant}).
Relative to the model's uncertainty, the effect is largest at level 1, where the model is
otherwise nearly certain and the command decides.

\subsection{Training details}
\label{sec:supp_training}
Our camera adapter is trained on 1,850 RealEstate10K~\cite{re10k} clips, 925 rotation-dominant and
925 translation-dominant, with camera poses estimated by MegaSAM~\cite{megasam}, for 23k steps, using
AdamW with learning rate $10^{-4}$ on the adapter only, on two A100 GPUs with one sample per GPU
(batch size 2).

The subject branch is trained on SA-V~\cite{sam2} clips with human-annotated masks, with the base model
and our camera adapter frozen, in the same setting (two A100 GPUs, batch size 2, learning rate
$10^{-4}$).
It is first trained for 23k steps on clips with a still camera, to follow the subject's outline.
It is then trained for 16k further steps on clips that add a moving camera.

\subsection{Control branches}
\label{sec:supp_branches}
This section gives the details of the control branches of the main text (\cref{sec:method:recipe},
\cref{fig:arch}).

\paragraph{Camera branch.}
Each token is given the ray through its pixel from the camera of its latent frame, expressed in the
first frame's camera.
Each of 8 transformer blocks gets a small parallel self-attention branch (8 heads, width 512),
whose queries, keys and values are its own projections of the block's input; the ray encoder maps
the six ray channels to the offsets $\Delta Q$ and $\Delta K$ of the branch's queries and keys.
The branch's output projection starts at zero, and its output is added to the block's residual
stream, so the base model's weights and its own attention are untouched.
The direction carries rotation and the moment carries translation; with the moment zeroed, the
translation error rises $7\times$ (\cref{tab:supp_moment}).

\begin{table}[t]
  \centering\small
  \begin{tabular}{@{}lcc@{}}
    \toprule
    Camera adapter, ray input & RotErr ($^\circ$) $\downarrow$ & TransErr $\downarrow$\\
    \midrule
    Direction only (moment zeroed) & 5.55 & 1.216\\
    Direction + moment             & \textbf{5.43} & \textbf{0.172}\\
    \bottomrule
  \end{tabular}
  \caption{\textbf{The ray moment carries translation: zeroing it raises the translation error
  $7\times$, with rotation almost unchanged.} A camera adapter with and without its moment input, on 40 held-out natural RealEstate10K~\cite{re10k} clips; errors re-measured by MegaSAM~\cite{megasam}.}
  \label{tab:supp_moment}
\end{table}

\paragraph{Subject branch.}
For a moving subject we add a second, gated branch of the same form that reads the subject's
moving outline (\cref{fig:arch}); its output gate starts at zero, so with no subject input the
model is exactly the camera model.

\section{Camera Control Experiments}
\label{sec:supp_experiments}
\label{sec:supp_camera}

\subsection{Camera control: detailed results}
\label{sec:supp_camera_details}
This section gives the full results of the camera experiments of the main text (\cref{sec:exp:camera}).
\textbf{Video results} for this section are on the \projectpage{} (``Camera control'' tab),
which shows our model and the baselines side by side with the commanded camera path.

\paragraph{Protocol.}
\emph{Commanded motions (test100):} Randomly sampled start frames of held-out RealEstate10K~\cite{re10k} scenes,
each driven along one of nine commanded paths (pans of $90^\circ$ right and left and of $45^\circ$, tilts of
$90^\circ$ and $30^\circ$, out and back; a strong zoom; pedestal, dolly and a $45^\circ$ orbit), 100 items in all
(\cref{tab:supp_camera_test100}). The dataset will be made public with the project code base.
We re-measure the generated camera with MegaSAM~\cite{megasam}; TransErr is scale-aligned to
RealEstate10K units.
Every system gets the same commanded path, and its generated camera is scored against that path
resampled to its own frame count.
test100 has no MegaSAM floor, since no real video follows its commanded paths; the floor measured on
real RealEstate10K test-split footage is given in \cref{sec:supp_camera_testsplit}.

\paragraph{Baseline settings.}
No camera-control design existed for a discrete next-scale video model other than RAYNOVA, so the
baselines necessarily differ from ours in backbone: all but RAYNOVA are continuous-latent diffusion
models.
Our model keeps the InfinityStar 8B backbone frozen and trains only the camera adapter
(\cref{sec:supp_training}).
Most of the baselines are trained on \emph{RealEstate10K, which our camera adapter is also trained on}:
CameraCtrl, MotionCtrl and AC3D, and RAYNOVA after our fine-tuning.

CameraCtrl~\cite{cameractrl} and MotionCtrl~\cite{motionctrl} add camera control to the image-to-video
Stable Video Diffusion model and are trained on RealEstate10K; we use these released versions.
AC3D~\cite{ac3d} adds a camera ControlNet to the CogVideoX-5B video diffusion transformer and is trained
on RealEstate10K; its released version is text-to-video, so we run it with a neutral indoor prompt, and only its camera is comparable.
DaS~\cite{das} fine-tunes the CogVideoX-5B image-to-video model to follow 3D point tracks rendered from the
first frame, and is trained on real videos from MiraData and synthetic Mixamo renderings, not on
RealEstate10K.

RAYNOVA~\cite{raynova} builds on the Infinity-2B image backbone~\cite{infinity} and was trained on driving video; we
fine-tune its released model end to end on our RealEstate10K training clips at the resolution of its own
reported training settings (constant learning rate $6.25{\times}10^{-6}$, from its own
learning-rate rule), for 42k samples, and report this checkpoint.
To check that it had plateaued, we continued training to 60k samples.
On test100 the camera error is unchanged (rotation and zoom $35.32^\circ$ vs.\ $36.50^\circ$, $p{=}0.63$;
translation $1.32$ vs.\ $1.44$, $p{=}0.29$).
On the RealEstate10K test split the rotation error is unchanged ($22.53^\circ$ vs.\ $23.38^\circ$,
$p{=}0.23$) and the translation error stays about four times ours ($1.23$ vs.\ $1.35$).
We discuss RAYNOVA, and why it is the only discrete next-scale model among our baselines, in
\cref{sec:supp_other_discrete}.

\paragraph{Commanded motions (test100).}
\Cref{tab:supp_camera_test100} breaks the test100 rows of main-text \cref{tab:camera_main} down by motion.
Our model, with the adapter at every level, has the lowest error on every motion but tilt $30^\circ$,
where AC3D is lower.
Over the rotation motions (the Rotations row of main-text \cref{tab:camera_main}), its error is
$11.10^\circ$: half that of CameraCtrl ($22.46^\circ$) and about a third of that of MotionCtrl and DaS
($30.34^\circ$ and $30.17^\circ$); it is lower than each of the three on $89$--$100\%$ of the items
(paired Wilcoxon $p{<}10^{-11}$ against each).
The margins are largest on the $90^\circ$ pans ($13.0^\circ$ and $9.7^\circ$ vs.\ $45.0^\circ$ and
$18.2^\circ$ for CameraCtrl) and the $90^\circ$ tilt.

On the translation motions, its error is half that of CameraCtrl ($0.264$ vs.\ $0.542$,
$p{=}5{\times}10^{-5}$). It is also lower than that of DaS on average ($0.585$), though not
significantly ($p{=}0.1$).

MotionCtrl turns the commanded way but stops at 7--14\% of the commanded rotation, and our error is
lower on every rotation item ($p{<}10^{-12}$).
On translations it moves much less than ours, and its error is $2.9{\times}$ ours ($0.767$ vs.\ $0.264$,
$p{<}10^{-4}$).

AC3D, although it does not see the start frame, is the strongest baseline on rotation, with
$15.10^\circ$ over the rotation motions against our $11.10^\circ$ ($p{=}2{\times}10^{-4}$).
Our error is lower on every motion except tilt $30^\circ$, where AC3D is lower.
AC3D, however, reaches less of the commanded rotation (median $62\%$, ours $75\%$), and its translation
error is $1.7{\times}$ ours ($0.441$ vs.\ $0.264$, $p{<}10^{-5}$).

\begin{table*}[t]
  \centering\small
  \setlength{\tabcolsep}{4.5pt}
  \begin{tabular*}{\linewidth}{@{\extracolsep{\fill}}lcccccccc@{}}
    \toprule
    & \multicolumn{2}{c}{\textbf{Ours}} & & & & & &\\
    \cmidrule(lr){2-3}
    Motion\hfill\emph{levels} & all & 1--3 & AC3D$^\dagger$~\cite{ac3d} & CameraCtrl~\cite{cameractrl} & MotionCtrl~\cite{motionctrl} & DaS~\cite{das} & RAYNOVA~\cite{raynova} & InfinityStar$^\ddagger$~\cite{infinitystar}\\
    \midrule
    \multicolumn{9}{@{}l}{\emph{Rotation and zoom motions} (RotErr $\downarrow$, $^\circ$)}\\
    pan $90^\circ$ right & \textbf{13.01} & 18.02 & 25.80 & 45.03 & 38.31 & 42.85 & 45.58 & 36.59\\
    pan $90^\circ$ left & \textbf{9.69} & 11.95 & 15.09 & 18.15 & 40.17 & 32.74 & 46.90 & 33.01\\
    pan $45^\circ$ & \textbf{5.24} & 8.51 & 7.52 & 10.17 & 18.63 & 13.04 & 24.16 & 18.21\\
    tilt $90^\circ$ & \textbf{21.93} & 22.93 & 22.52 & 29.44 & 38.63 & 44.60 & 48.03 & 47.24\\
    tilt $30^\circ$ & 3.68 & 4.48 & \textbf{2.04} & 5.39 & 12.05 & 11.93 & 20.70 & 19.76\\
    strong zoom & \textbf{1.11} & 2.25 & 2.26 & 1.74 & 2.85 & 5.81 & 15.55 & 4.54\\
    \midrule
    all rotations & \textbf{11.10} & 13.51 & 15.10 & 22.46 & 30.34 & 30.17 & 38.00 & 31.87\\
    all rotation and zoom & \textbf{10.43} & 12.76 & 14.24 & 21.08 & 28.51 & 28.55 & 36.50 & 30.05\\
    \midrule
    \multicolumn{9}{@{}l}{\emph{Translation motions} (TransErr $\downarrow$)}\\
    pedestal & \textbf{0.371} & 0.397 & 0.662 & 0.743 & 0.421 & 0.473 & 1.549 & 1.563\\
    dolly & \textbf{0.178} & 0.233 & 0.315 & 0.319 & 0.892 & 0.299 & 1.374 & 0.838\\
    orbit $45^\circ$ & \textbf{0.222} & 0.273 & 0.253 & 0.587 & 1.207 & 1.381 & 1.376 & 1.436\\
    \midrule
    all translation & \textbf{0.264} & 0.307 & 0.441 & 0.542 & 0.767 & 0.585 & 1.445 & 1.247\\
    \midrule
    Min.\ per 10\,s video\,$\downarrow$ & \multicolumn{2}{c}{\textbf{2.3}} & 7.9 & 3.9 & 3.6 & 12.1 & 8.2 & \textbf{2.3}\\
    FVD\,$\downarrow$ & \textbf{591} & 620 & 1092 & 743 & 932 & 716 & 1108 & 643\\
    \bottomrule
  \end{tabular*}
  \caption{\textbf{Ours has the lowest camera error on almost every commanded motion, with the
  lowest generation time and FVD.} Camera control on test100, per motion. Errors of the generated camera, re-measured by MegaSAM~\cite{megasam}, with the reference path resampled to each
  system's frame count.
  Ours: camera adapter at all levels, or only at levels 1--3.
  $^\ddagger$InfinityStar without camera control: the same backbone with our adapter switched off, given the same
  start frame, seed and caption, which names the commanded motion.
  $^\dagger$AC3D is released only as text-to-video: it does not see the start frame, so only its camera is comparable.
  The camera paths go out and back, so a model that barely moves also has a moderate rotation
  error: DaS and MotionCtrl reach only 1--30\% and 7--14\% of the commanded rotation, against
  47--86\% for ours. MotionCtrl has no intrinsics input, so its zoom entry measures only stray rotation.
  Last rows: minutes per 10\,s video on one A100 (model loading excluded) and FVD against 80 held-out
  RealEstate10K clips (I3D, every video resampled to 16 frames evenly spaced in time).}
  \label{tab:supp_camera_test100}
\end{table*}

\paragraph{Visual quality and speed on test100.}
The last two rows of \cref{tab:supp_camera_test100} give the generation time per 10\,s video and FVD.
For FVD, every video, generated or real, is resampled to 16 frames evenly spaced over its duration, so
that the systems' different native frame counts do not affect the score; frames are
resized to a short side of 224 pixels and centre-cropped, and their I3D features are compared with
those of 80 held-out RealEstate10K clips.
Our videos have the lowest FVD ($591$; the next lowest is DaS at $716$).
AC3D generates its scenes from text, so its FVD against RealEstate10K ($1092$) also measures how far its
rooms are from real ones.

Our model generates 10\,s of video in 2.3 minutes, against 3.6 for MotionCtrl, 3.9 for CameraCtrl,
7.9 for AC3D, 8.2 for RAYNOVA and 12.1 for DaS (\cref{tab:supp_camera_test100}), on one A100 80GB with
model loading excluded.
Each model runs at its native resolution and frame rate (ours $848{\times}480$, 161 frames; CameraCtrl
$576{\times}320$, $4{\times}25$ frames; MotionCtrl $1024{\times}576$, $7{\times}14$ frames; DaS and AC3D
$720{\times}480$, $2{\times}49$ frames; RAYNOVA 97 frames in a sliding window, at its training
resolution); models without a long-video mode chain their native chunks.

\subsection{Camera control on the RealEstate10K test split}
\label{sec:supp_camera_testsplit}
Our model is most accurate on the user-commanded motions of test100 (\cref{sec:supp_camera_details}).
To check that it is also competitive on a standard camera-control benchmark, we evaluate on the official
RealEstate10K test split~\cite{re10k}, which no system was trained on, with each clip's own camera path
as the command (299 clips).
The generated camera is re-measured by MegaSAM~\cite{megasam} and scored against the clip's pose file
resampled to each system's frame count.
As a reference for the measurement itself, MegaSAM re-measuring the real clips against their pose files
errs by $0.97^\circ$ / $0.117$ (the \emph{MegaSAM floor}).
We encourage readers to view four of these clips, each next to its recording, on the \projectpage{}
(``Camera control'' tab, ``RealEstate Test Clips'').

\begin{table}[t]
  \centering\small
  \setlength{\tabcolsep}{4pt}
  \begin{tabular}{@{}lcc@{}}
    \toprule
    RealEstate10K test split (299 clips) & RotErr\,$\downarrow$ & TransErr\,$\downarrow$\\
    \midrule
    \rowcolor{refgrey} MegaSAM floor (real video) & 0.97 & 0.117\\
    \midrule
    InfinityStar~\cite{infinitystar}, no camera & 19.48 & 0.971\\
    RAYNOVA~\cite{raynova} & 23.38 & 1.350\\
    DaS~\cite{das} & 11.08 & 0.556\\
    MotionCtrl~\cite{motionctrl} & 14.91 & 0.706\\
    CameraCtrl~\cite{cameractrl} & 5.00 & 0.342\\
    AC3D$^\dagger$~\cite{ac3d} & 4.61 & 0.339\\
    \midrule
    Ours (levels 1--3)                       & 4.64 & 0.344\\
    \textbf{Ours (all levels)} & \textbf{4.56} & \textbf{0.314}\\
    \bottomrule
  \end{tabular}
  \caption{\textbf{On real camera paths that no system trained on, ours has the lowest rotation and
  translation error.} RealEstate10K test split, 299 clips. Mean errors of the generated camera, re-measured by MegaSAM~\cite{megasam}, with the clip's own path
  as the command. RotErr in degrees; lower is
  better for every column. MegaSAM floor: MegaSAM re-measuring the real clips.
  $^\dagger$AC3D is text-to-video and does not see the first frame.}
  \label{tab:camera_testsplit}
\end{table}

On these clips our model has the lowest mean error in both rotation and translation
(\cref{tab:camera_testsplit}), narrowly ahead of AC3D and CameraCtrl ($4.56^\circ$ vs.\ $4.61^\circ$ and
$5.00^\circ$), so it is competitive on standard camera paths as well; its margin grows on the larger
commanded motions of test100.

\subsection{Does placing the control signal in the control band matter?}
\label{sec:supp_placement}
The main text (\cref{sec:exp:placement}) trains camera adapters that differ in where the conditioning
acts. Here we report all of them: (A$'$) levels 1--2, the core of the band; (A) levels 1--3, the control
band; (B) every level, and (C) levels 5--13, the render
bands where motion becomes visible.
Outside its levels, an arm's adapter is switched off, so no ray direction or moment reaches them.
We evaluate on 40 held-out natural RealEstate10K clips (\cref{tab:supp_placement}).

\begin{table}[t]
  \centering\small
  \setlength{\tabcolsep}{4pt}
  \begin{tabular}{@{}lcc@{}}
    \toprule
    Camera adapter active at & RotErr ($^\circ$) $\downarrow$ & TransErr $\downarrow$\\
    \midrule
    (A$'$) levels 1--2 ($0.36\%$)             & 7.48 & 0.236\\
    (A) levels 1--3, control band ($0.7\%$)  & 6.95 & 0.270\\
    (B) every level                           & \textbf{6.69} & \textbf{0.235}\\
    (C) levels 5--13, render bands ($98.5\%$) & 18.41 & 0.812\\
    No camera (adapter off)                   & 18.02 & 0.725\\
    \bottomrule
  \end{tabular}
  \caption{\textbf{Conditioning $0.7\%$ of the tokens (A, the control band) recovers $98\%$ of the
  every-level rotation gain and $93\%$ of its translation gain.} Gain: improvement over no camera, relative
  to (B). Conditioning the render bands (C, $98.5\%$ of the tokens) is no better than no camera.
  Percentages: share of the clip's tokens. 40 held-out natural RealEstate10K clips~\cite{re10k}, errors
  re-measured by MegaSAM~\cite{megasam} with the reference path resampled to each system's frame count.}
  \label{tab:supp_placement}
\end{table}

\textbf{Conditioning $0.7\%$ of the tokens is enough on these natural paths.}
The core of the band alone, (A$'$), conditions $0.36\%$ of the tokens and already recovers $93\%$ of B's
rotation gain over no camera and all of its translation gain ($7.48^\circ$ and $0.236$ vs.\ $6.69^\circ$ and
$0.235$).
With the whole control band, (A) conditions $0.7\%$ of the tokens and recovers $98\%$ of B's rotation gain
and $93\%$ of its translation gain; (A) and B are not significantly different ($p{=}0.57$), although B
conditions every token.
On the larger commanded motions of test100, B is more accurate than (A), by $1.25^\circ$ in rotation.

\textbf{Control placed after the band comes too late.}
C is no better than no camera ($18.41^\circ$ vs.\ $18.02^\circ$), and every arm with control in the band is
far better than C on most clips.
Why C fails is visible in its tokens.
Its command first enters at level 5, so for any one scene C generates identical tokens at levels
0--4 whatever the command, and all seven test commands, from ``stay still'' to a $45^\circ$
pan, give the same camera motion.
The command does reach the fine levels: it shifts their predicted code distributions as much
as in the full model.
By then the trajectory is committed, and nothing is left for the command to steer: control must act
before motion appears at level 5 (main text, \cref{sec:method:recipe}).

\textbf{The same holds within the weights of our model.}
We take our model, trained with the adapter at every level, and switch the adapter off outside a chosen
set of levels at inference only (same weights, seed and first frame), on the 100 test100 items
(\cref{tab:supp_gating}).
\begin{table}[t]
  \centering\small
  \setlength{\tabcolsep}{3pt}
  \begin{tabular}{@{}lccccc@{}}
    \toprule
    Adapter at & Tokens & RotErr\,$\downarrow$ & kept & TransErr\,$\downarrow$ & kept\\
    \midrule
    Every level   & $100\%$  & \textbf{10.43} & 100\% & \textbf{0.264} & 100\%\\
    Levels 1--3   & $0.7\%$  & 12.76 & 88\% & 0.307 & 96\%\\
    Levels 1--2   & $0.36\%$ & 14.46 & 79\% & 0.397 & 86\%\\
    Levels 3--4   & $1.1\%$  & 27.45 & 13\% & 1.167 & 8\%\\
    Level 4       & $0.7\%$  & 30.10 & 0\%  & 1.215 & 3\%\\
    Levels 5--13  & $98.5\%$ & 29.43 & 3\%  & 1.213 & 3\%\\
    None (no camera) & $0\%$ & 30.05 & -- & 1.247 & --\\
    \midrule
    Levels 1--3$^\ast$ & $0.7\%$ & 11.02 & 97\% & 0.351 & 91\%\\
    \bottomrule
  \end{tabular}
  \caption{\textbf{In the same weights, the camera acts through levels 1--3: the adapter there alone keeps
  $88\%$ of the rotation gain and $96\%$ of the translation gain; at levels 5--13 it keeps almost none.}
  Inference-time gating of our model on test100 (RotErr in degrees over the rotation and zoom
  items, TransErr over the translation items). ``Kept'': share of the every-level
  model's improvement over no camera that remains. Tokens: share of all tokens in the clip's full pyramid
  (levels 0--13, 77{,}980 tokens; \cref{tab:levels}) that the adapter conditions.
  $^\ast$Trained with the adapter at levels 1--3 only (same recipe), and tested the same way.}
  \label{tab:supp_gating}
\end{table}
With the adapter only at levels 1--3, the model keeps $88\%$ of its rotation gain over no camera and $96\%$
of its translation gain; levels 1--2 alone keep $79\%$ and $86\%$, so level 3 adds to them.
Any set without levels 1--2 is close to no camera: level 4 alone changes nothing, levels 3--4 recover
about a tenth, and levels 5--13 recover none.
Trained with the adapter at levels 1--3 only, the same recipe recovers most of the rotation gap to every level
($97\%$ of the gain, against $88\%$ when the adapter is switched off at inference), while translation stays
a little behind ($91\%$).

\textbf{Where to place the control is therefore a choice by task}, as in the main text: the control band alone
($0.7\%$ of the tokens) gives nearly full accuracy on natural camera paths (\cref{tab:supp_placement}),
while control at every level gives the most accurate large commanded motions
(\cref{tab:supp_gating,tab:supp_camera_test100}), and the camera results of the main text use every level.
Even for these large motions, the decision is made in the band: copying a donor's levels $0..3$ transfers
its camera motion also for the largest commands (\cref{tab:supp_transplant}).

\subsection{Other discrete next-scale video models}
\label{sec:supp_other_discrete}

\paragraph{RAYNOVA is the closest released model.}
Our recipe needs a next-scale video model with discrete tokens.
InfinityStar~\cite{infinitystar} is, to our knowledge, the only such model released for general,
open-domain video, so it is the backbone of our model.
The closest released model is RAYNOVA~\cite{raynova}, which extends the next-scale image model
Infinity-2B~\cite{infinity} to multi-camera driving video, one frame at a time, and is trained on driving
data (nuPlan and nuScenes).
Other discrete next-scale video models have been described but not released, so they cannot be run:
VideoAR~\cite{videoar} and ERNIE~5.0~\cite{ernie5}, which predict scales within each frame and frames
in sequence, and WALL-SS~\cite{wallss}, a robot world model initialized from InfinityStar.
There is therefore no other discrete next-scale model with camera control on which a same-backbone
comparison could be run.

\paragraph{Its design and data explain its weak scores in our camera tables.}
RAYNOVA conditions on camera rays, but it gives no recipe for where camera control should enter the
model and does not report how accurately the commanded camera is followed.
Its backbone is an earlier and smaller image model: it builds a separate 7-scale pyramid for each
frame and makes video by conditioning each frame on a short window of previous frames, whereas
InfinityStar builds one pyramid for the whole clip with a video tokenizer.
Its purpose and data also differ from ours: as released, it turns any indoor first frame into a street
scene from the first generated frame, so it cannot be scored on our indoor benchmarks.
Even after fine-tuning on our data, its videos are of lower quality than those of InfinityStar
without any camera input (FVD 1108 vs.\ 643, \cref{tab:supp_camera_test100}), and it follows the
commanded camera less accurately than InfinityStar given no camera at all (rotation and zoom
error $36.50^\circ$ vs.\ $30.05^\circ$ on test100, rotation error $23.38^\circ$ vs.\ $19.48^\circ$ on the test split;
\cref{tab:supp_camera_test100,tab:camera_testsplit}).
These differences, rather than next-scale prediction itself, likely account for its low place in our camera tables.

\paragraph{We repeat our hierarchy analyses on RAYNOVA.}
We use the RAYNOVA model fine-tuned on our RealEstate10K training clips, the one in our camera tables
(Baseline settings, \cref{sec:supp_camera_details}).
At its training resolution each frame has 7 scales, $0$--$6$, with 472 tokens, so we apply each of
our analyses (\cref{sec:supp_bands,sec:supp_control}) per frame, on the same clips, scenes and
commands as for our model.
Four findings carry over.

\paragraph{1. Motion is rendered in the middle of the pyramid, as in ours.}
Partial decodes of RAYNOVA's tokenizer on our 80 held-out clips, scored with the progress rule of
\cref{tab:supp_render}, show no motion at scales $0$--$2$.
Colour appears first (scale 2), then motion together with coarse layout (scale 3), then identity
and edges (scale 4).
As in ours, no motion is visible at the coarsest scales; it appears in the middle of the pyramid.

\paragraph{2. The command enters at the same scale.}
With the logit probe of \cref{sec:supp_control} on the same 8 scenes, the command never changes the
prediction at scale 0 ($|\Delta p| = 0$) and changes it most at scale 1, both in absolute terms
($0.028$, about twice the value at scales 2--6) and relative to the model's uncertainty, as at level 1
of our model.

\paragraph{3. The camera is set before the finest scales.}
With the camera transplant on the same 20 held-out scenes and command pairs
(\cref{tab:supp_transplant}), copying a donor's scales $0$--$4$ into every generated frame (108 of
472 tokens, $23\%$) transfers its camera motion: median transfer $0.92$ for standard commands and
$0.83$ for large ones, and 19 of 20 calls told to pan the opposite way follow the donor.
Unlike in our model, which reaches $0.96$ at level 2, transfer builds up over scales 1--4
($0.23$, $0.58$, $0.69$, $0.92$).

\paragraph{4. The finest scales do not need the camera.}
We fine-tune RAYNOVA again from its released weights with the same recipe, but give it the real
camera only at scales $0$--$4$ and a still camera at scales 5--6.
It follows the camera as well as the RAYNOVA model of our camera tables: rotation and zoom error
$38.75^\circ$ vs.\ $36.50^\circ$ on test100 ($p{=}0.47$) and $23.79^\circ$ vs.\ $23.38^\circ$ on the
test split ($p{=}0.84$); translation error $1.46$ vs.\ $1.44$ ($p{=}0.43$) and $1.33$ vs.\ $1.35$
($p{=}0.46$).

\paragraph{RAYNOVA's weak camera following limits all of these comparisons.}
RAYNOVA barely follows the camera in any of these runs, with rotation and zoom errors of
$36$--$40^\circ$ on test100 against our $10.43^\circ$, so each difference above is a few degrees.
This weakness confounds both its camera-control results and its band analyses.
Our gating test (\cref{tab:supp_gating}) is uninformative on RAYNOVA for this reason: the RAYNOVA model of our
camera tables is no more accurate than with a still camera.
The analyses above therefore show where RAYNOVA's camera information sits, not how much control each
placement gives.
Making RAYNOVA follow the camera closely would likely require a new architecture design, which is
beyond the scope of this work.

\subsection{Visual comparison}
\label{sec:supp_camera_visual}
\begin{figure*}[p]
  \centering
  \includegraphics[width=\linewidth]{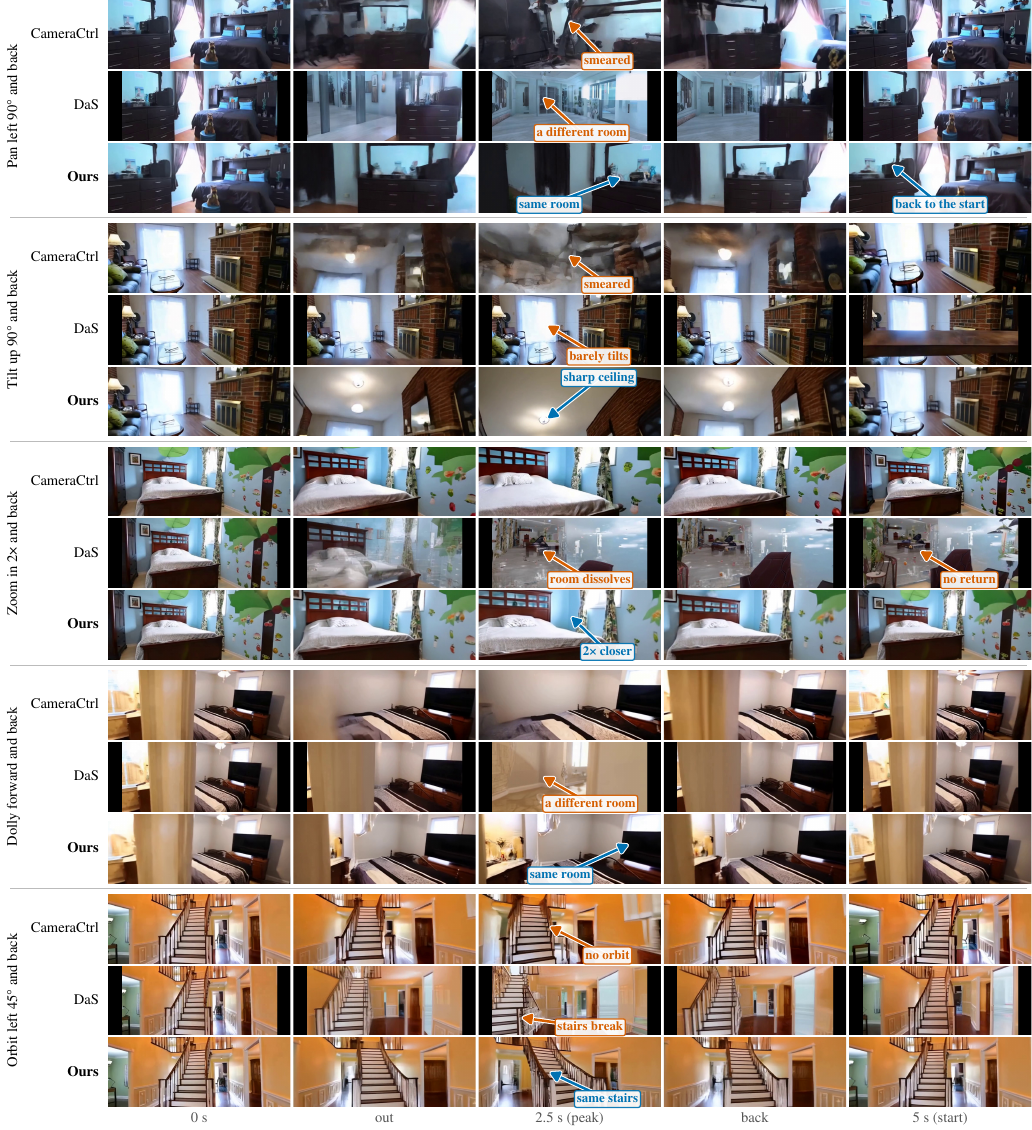}
  \caption{\textbf{Five commanded motions, out and back in 5\,s.} Held-out RealEstate10K~\cite{re10k} start frames;
  frames at 0\,s, halfway out, the peak of the motion (2.5\,s), halfway back and the end. Orange: where a baseline
  smears the view or loses the scene; blue: our model. Examples from the \projectpage.}
  \label{fig:supp_camera_gallery}
\end{figure*}

\Cref{fig:supp_camera_gallery} extends main-text \cref{fig:camera} to five more commanded motions of test100, against the
two baselines of the main text's comparison. As the camera moves away from the start frame, CameraCtrl smears the new
views of the large rotations, and DaS barely moves or replaces the scene with a different one; our model shows a sharp
view of the same scene and returns to the start. The videos of these and further examples, with all baselines, are on the \projectpage{} (``Camera control'' tab).

\clearpage

\section{Subject Control Experiments}
\label{sec:supp_subject}

\begin{figure*}[t]
  \centering
  \includegraphics[width=\linewidth]{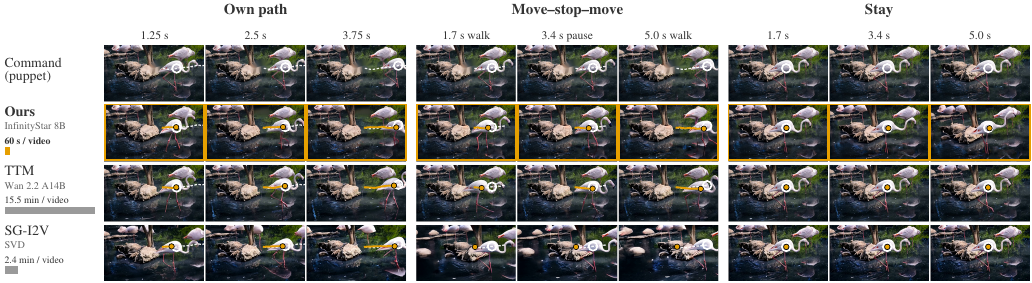}
  \caption{\textbf{One flamingo, three commands.} DAVIS~\cite{davis2017} \emph{flamingo}, still
  camera. White dashes: the commanded path; white ring: where the command puts the subject at
  that time; orange: the tracked subject so far; bars: time to generate one 5\,s video on one A100.
  On its own path, all three methods walk the flamingo (error 0.05, 0.05 and 0.07).
  On move--stop--move, ours stands still during the pause and then walks on; TTM keeps walking
  through the pause (pause motion $0.23$ of the path) and does not walk on, and SG-I2V leaves the
  flamingo in place while the background shifts.
  Under stay, all three hold the flamingo still.}
  \label{fig:supp_subject_compare}
\end{figure*}

\begin{table}[t]
  \centering
  \setlength{\tabcolsep}{2.5pt}
  \resizebox{\linewidth}{!}{%
  \begin{tabular}{@{}lccccccc@{}}
    \toprule
    & \multicolumn{2}{c}{Own path} & \multicolumn{3}{c}{Move--stop--move} & Stay & Time\\
    \cmidrule(lr){2-3}\cmidrule(lr){4-6}\cmidrule(lr){7-7}\cmidrule(l){8-8}
    Method & Err.\,$\downarrow$ & Prog. & Err.\,$\downarrow$ & Pause\,$\downarrow$ & Restart\,$\uparrow$ & Held\,$\uparrow$ & (s)\,$\downarrow$\\
    \midrule
    \multicolumn{8}{@{}l}{\emph{No command}}\\
    InfinityStar 8B, no adapter~\cite{infinitystar} & 0.24 & 40\% & -- & -- & -- & 1/14 & 60\\
    Our camera model           & 0.24 & 59\% & -- & -- & -- & 0/14 & 60\\
    Wan 2.2 I2V~\cite{wan}     & 0.21 & 57\% & -- & -- & -- & 0/14 & 960\\
    \midrule
    \multicolumn{8}{@{}l}{\emph{Training-free subject control}}\\
    TTM~\cite{ttm}, Wan 2.2 A14B   & \textbf{0.05} & 89\% & 0.08 & 0.09 & 8/14 & \textbf{11/14} & 931\\
    TTM~\cite{ttm}, SVD-xt         & 0.35 & 34\% & 0.16 & 0.12 & 6/14 & \textbf{11/14} & 50\\
    SG-I2V~\cite{sgi2v}, SVD       & 0.36 & 24\% & 0.27 & \textbf{0.02}$^\dagger$ & 7/14 & 7/14 & 142\\
    \midrule
    \multicolumn{8}{@{}l}{\emph{Ours} (InfinityStar 8B)}\\
    Ours: subject branch only     & 0.11 & 87\% & 0.11 & 0.23 & 7/14 & 0/14 & 60\\
    Ours: whole-frame transplant (no training) & 0.07 & 89\% & 0.07 & 0.05 & 11/14 & \textbf{11/14} & 60\\
    Ours: region transplant (no training) & 0.07 & 88\% & 0.06 & 0.05 & \textbf{12/14} & 8/14 & 60\\
    \textbf{Ours: region transplant + subject branch} & 0.06 & 87\% & \textbf{0.04} & 0.04 & \textbf{12/14} & \textbf{11/14} & 60\\
    \midrule
    \rowcolor{refgrey} Ceiling: real clip's levels 0--3 & 0.06 & 89\% & -- & -- & -- & -- & 60\\
    \bottomrule
  \end{tabular}}
  \caption{\textbf{Our path error matches the ceiling and TTM on Wan 2.2 at $15\times$ faster generation,
  and our subject stops and restarts better.} Subject control, still camera, 14 DAVIS~\cite{davis2017}
  sequences. Medians over sequences; metrics as defined at the start of \cref{sec:supp_subject}.
  $^\dagger$SG-I2V barely moves the subject at all (move--stop--move error 0.27).}
  \label{tab:subject}
\end{table}

\begin{table*}[t]
  \centering\small
  \setlength{\tabcolsep}{4pt}
  \begin{tabular}{@{}lccccccccr@{}}
    \toprule
    & \multicolumn{2}{c}{Own path (80)} & \multicolumn{4}{c}{Move--stop--move (80)} & \multicolumn{2}{c}{Stay (90)} & \\
    \cmidrule(lr){2-3}\cmidrule(lr){4-7}\cmidrule(lr){8-9}
    Method & Err.\,$\downarrow$ & Scale\,$\rightarrow$1 & Err.\,$\downarrow$ & Pause\,$\downarrow$ & Pauses\,$\uparrow$ & Moves again\,$\uparrow$ & Drift\,$\downarrow$ & Held\,$\uparrow$ & Time (s)\\
    \midrule
    \multicolumn{10}{@{}l}{\emph{No command}}\\
    InfinityStar 8B, no adapter~\cite{infinitystar} & 0.39 & 0.26 & -- & -- & -- & -- & 0.187 & 7/90 & 60\\
    Our camera model              & 0.43 & 0.04 & -- & -- & -- & -- & 0.155 & 21/90 & 60\\
    Wan 2.2 I2V~\cite{wan}        & 0.37 & 0.54 & -- & -- & -- & -- & 0.349 & 5/90 & 960\\
    SVD-xt~\cite{svd}             & 0.48 & 0.01 & -- & -- & -- & -- & 0.016 & 48/90 & 71\\
    SVD~\cite{svd}                & 0.45 & 0.04 & -- & -- & -- & -- & 0.059 & 27/90 & 27\\
    \midrule
    \multicolumn{10}{@{}l}{\emph{Training-free subject control}}\\
    TTM~\cite{ttm}, Wan 2.2 A14B  & 0.18 & 0.74 & 0.09 & 0.04 & 59/80 & 47/80 & 0.003 & 72/90 & 931\\
    TTM~\cite{ttm}, SVD-xt        & 0.40 & 0.06 & 0.20 & 0.02 & 46/80 & 35/80 & 0.003 & 71/90 & 50\\
    SG-I2V~\cite{sgi2v}, SVD      & 0.42 & 0.05 & 0.26 & 0.01 & 47/80 & 45/80 & 0.008 & 55/90 & 142\\
    \midrule
    \multicolumn{10}{@{}l}{\emph{Ours} (InfinityStar 8B)}\\
    Ours: subject branch only     & 0.31 & 0.44 & 0.21 & 0.07 & 51/80 & 46/80 & 0.107 & 15/90 & 60\\
    Ours: whole-frame transplant (no training) & \textbf{0.15} & 0.75 & 0.10 & \textbf{0.00} & \textbf{63/80} & \textbf{62/80} & \textbf{0.002} & \textbf{75/90} & 60\\
    \textbf{Ours: region transplant + subject branch} & \textbf{0.15} & 0.76 & \textbf{0.08} & 0.01 & 62/80 & 58/80 & 0.008 & 61/90 & 60\\
    \midrule
    \rowcolor{refgrey} Ceiling: real clip's levels 0--3 & 0.17 & 0.79 & -- & -- & -- & -- & -- & -- & 60\\
    \bottomrule
  \end{tabular}
  \caption{\textbf{As on the 14 sequences of the main text (its \cref{tab:subject_still}; \cref{tab:subject} here), on all of DAVIS our region transplant
  matches TTM on Wan 2.2 in path error and pauses and restarts more cleanly, at $15\times$ faster generation.} Subject control on all of DAVIS~\cite{davis2017}, still
  camera, pre-registered subsets: the 80 sequences with a moving subject for own path and move--stop--move, all 90
  sequences for stay. Medians over sequences; metrics as defined at the start of \cref{sec:supp_subject}.
  Copying the whole frame also copies the camera (main text, \cref{sec:method:recipe}), so only our region
  transplant can be combined with camera control.
  Against InfinityStar without an adapter, our region transplant follows the path more closely and drifts
  less under stay in most sequences.}
  \label{tab:supp_davis_all}
\end{table*}

\paragraph{Metrics:}
As in the main text (\cref{sec:exp:subject}), we track the subject with SAM\,2~\cite{sam2} and remove
camera motion with background homographies; distances are in frame widths.

\textbf{With a still camera}, \emph{Err.} is the mean distance between the tracked and the commanded path
(ObjMC~\cite{motionctrl}), divided by the length of the real path; \emph{Prog.} is how far along the
real path the subject ends; \emph{Scale} is the least-squares factor between the generated and the
commanded path ($1$ = right distance).
\emph{Pause} is how far along the path the subject moves during the commanded pause; \emph{Pauses} counts
the sequences in which it moves less during the pause than in its own-path video, and \emph{Restart}
(\emph{Moves again}) those in which it moves more after the pause than during it.
\emph{Drift} is the subject's largest shift from where it stood in frame 0 under stay; \emph{Held} counts
the stay videos whose drift stays within $2\%$ of the frame width.
\emph{Frame change} is the mean absolute pixel change between frames 8 apart over the whole frame, relative
to our camera model without a subject command on the same sequence; it falls when the scene stops moving.

\textbf{With a moving camera}, \emph{Err.} is measured on the half and the full turn; \emph{Worse than still}
counts runs whose error exceeds the same method's still-camera error; \emph{Camera reach} is the measured
turn as a fraction of the commanded one; motion smoothness is VBench's~\cite{vbench}.
Time is per 5\,s video on one A100; paired comparisons are one-sided sign tests over sequences.

\subsection{Subject control: still camera}
\label{sec:supp_subject_all}

\Cref{tab:subject} gives the full subject-control results on the 14 DAVIS sequences of the main
text (\cref{sec:exp:subject} and \cref{tab:subject_still}).
We encourage readers to view the generated videos on the \projectpage{} (``Subject control'' tab,
``Still camera''), which shows move--stop--move commands for ours and the baselines.

\paragraph{Results.}
Without a command, neither our camera model nor Wan 2.2 follows the path or
holds the subject still (0/14; \cref{tab:subject}).

Our region transplant with the subject branch brings the error to 0.06 of the path length, the same as
the control ceiling that transplants the real clip's own levels 0--3 (0.06).
It matches its continuous diffusion counterpart, TTM on Wan 2.2 (0.05; the difference is not
significant), while generating a video $15\times$ faster (60\,s against 931\,s on one A100,
\cref{fig:supp_subject_compare}).
Against the baselines of comparable cost, TTM on SVD-xt and SG-I2V, our error is lower on 13/14 and
14/14 sequences.

\textbf{Our subject also keeps time better} than TTM on Wan 2.2 (\cref{fig:supp_subject_compare}).
Its move--stop--move error is half of TTM's (0.04 against 0.08, $p{=}0.03$), it stops more cleanly
during the pause (pause motion 0.04 against 0.09, $p{=}0.007$), and it walks on after the pause in 12/14
sequences against 8/14.
Under stay, both hold the subject still in 11/14 sequences.

\textbf{The transplant does most of the work of commanding the subject}, with no training. Written into the
coarse levels 0--3, it already reaches an error of 0.07, whether it covers the whole frame or only the subject's
region, while the subject branch alone follows only loosely (0.11) and neither stops nor stays (0/14 held).
\textbf{The branch is what makes the transplant usable with a moving camera.}
Copying the whole frame also copies the camera (main text, \cref{sec:method:recipe}), so only the region
transplant can be combined with camera control, and the region transplant alone holds fewer subjects under
stay (8/14) and, once the camera turns, rises to TTM's error (\cref{sec:supp_subject_pan}).
Adding the branch lowers the move--stop--move error from 0.06 to 0.04 and holds more subjects under stay (11/14
against 8/14).

\textbf{On all 80 DAVIS sequences with a moving subject (\cref{tab:supp_davis_all}), the main text's picture
holds}: our own-path error ties TTM on Wan 2.2 (0.15 against 0.18) and matches the ceiling (0.17), and our
subject pauses and restarts more cleanly than TTM's in most sequences ($p{\le}0.001$). The
move--stop--move error ties (0.08 against 0.09).
Both still-camera claims of the main text therefore extend from its 14 sequences to all of DAVIS: ours
matches TTM on the subject's own path at $15\times$ the speed, and it keeps better time through a pause,
by a wider margin here (pause motion a quarter of TTM's, against about half on the 14 sequences).
The whole-frame transplant matches or beats ours on most measures here because it copies the puppet's whole frame, whose
background is frame 0 held still: the rest of the scene stops moving (frame change 0.33, against 0.46 for
ours; ours higher in 69/80), and our pause and stay measures reward a still frame.
Under stay, TTM and the whole-frame transplant hold more subjects still, but both still the scene to do it
(frame change 0.22 and 0.17 against 0.31; ours livelier than TTM in 60/90).

\Cref{fig:supp_subject_compare} shows the three commands (own path, move--stop--move and stay)
on one DAVIS sequence.

\subsection{Subject control: moving camera}
\label{sec:supp_subject_pan}

\begin{table}[t]
  \centering\footnotesize
  \setlength{\tabcolsep}{2pt}
  \resizebox{\linewidth}{!}{%
  \begin{tabular}{@{}lcccc@{}}
    \toprule
    & \multicolumn{2}{c}{Err.\,$\downarrow$} & Worse than & Camera\\
    \cmidrule(lr){2-3}
    Method & half & full & still\,$\downarrow$ & reach\\
    \midrule
    TTM~\cite{ttm}, camera recipe & 0.10 & 0.12 & 20/28 & 0.97 / 0.97\\
    TTM~\cite{ttm}, cut-and-drag  & 0.09 & 0.10 & 20/28 & 0.94 / 0.93\\
    TTM, SVD-xt, camera mask      & 0.36 & 0.48 & 20/28 & 0.05 / 0.03\\
    TTM, SVD-xt, cut-and-drag     & 0.40 & 0.47 & 21/28 & 0.04 / 0.01\\
    SG-I2V~\cite{sgi2v}, + backgr.\ boxes & 0.32 & 0.47 & 17/28 & 0.38 / 0.20\\
    SG-I2V~\cite{sgi2v}, subject box & 0.50 & 0.51 & 25/28 & $-0.74^\ddagger$ / 0.08\\
    Ours: camera only             & 0.16 & 0.21 & 8/28$^\ast$ & 1.21 / 0.70\\
    Ours: subject branch only     & 0.14 & 0.11 & 15/28 & 1.20 / 0.83\\
    Ours: without subject branch  & 0.11 & 0.10 & 19/28 & 0.94 / 0.91\\
    \textbf{Ours: reg. transplant + subj. branch} & \textbf{0.04} & \textbf{0.07} & \textbf{8/28} & 0.98 / 0.95\\
    \bottomrule
  \end{tabular}}
  \caption{\textbf{As in the main text (its \cref{tab:subject_pan}), under a turning camera the subject error of
  our region transplant with the subject branch stays near its still-camera level while TTM's rises; without
  either part, ours rises to TTM's level too.}
  Subject control with a moving camera, own path, 14 DAVIS sequences; TTM on Wan 2.2 unless marked SVD-xt;
  metrics as defined at the start of \cref{sec:supp_subject}. The camera turns towards where the subject walks,
  by half or all of its sideways travel; camera reach is given for the half / full turn.
  Without subject branch: region transplant only. Camera only: neither transplant nor branch.
  $^\ast$Camera only has no subject command and does not follow the path even with a still camera (error 0.24),
  so it is rarely worse than that.
  $^\ddagger$SG-I2V's single box drags the whole scene with it, which reads as a turn the wrong way.}
  \label{tab:supp_subject_pan}
\end{table}

\begin{table}[t]
  \centering\footnotesize
  \begin{tabular}{@{}lcc@{}}
    \toprule
    Ours vs. & Half turn & Full turn\\
    \midrule
    TTM~\cite{ttm}, camera recipe & 9/14 & 11/14\\
    TTM~\cite{ttm}, cut-and-drag  & 11/14 & 10/14\\
    \bottomrule
  \end{tabular}
  \caption{\textbf{Our subject stays closer to its path than TTM's in most sequences, in all four
  comparisons.} The paired comparisons behind the main text's \cref{sec:exp:subject}: for each TTM recipe and
  camera turn, the number of the 14 sequences in which our subject error is lower than TTM's.}
  \label{tab:supp_subject_pan_paired}
\end{table}

On the same 14 sequences, each own-path command is repeated with the camera turning towards where
the subject walks: by half of its sideways travel (\emph{half}, $3$--$15^\circ$) or all of it
(\emph{full}, $3$--$45^\circ$).
We compare with TTM on Wan 2.2 in two recipes, its cut-and-drag recipe, as in the still-camera
experiments, and its camera recipe, and with the SVD-based baselines (\cref{tab:supp_subject_pan}).
We encourage readers to view the videos on the \projectpage{} (``Subject control'' tab,
``Moving camera'').

\textbf{Both follow the camera.} TTM on Wan 2.2 and our region transplant with the subject branch both turn
as commanded (camera reach 0.93--0.98 on both turns), while the SVD-based baselines barely turn (at most
0.20 of the full turn) and lose the subject's path (error 0.32--0.51).
The main text reports reach on the full turn only; the half turn shows the same.

\textbf{Our subject stays closer to its path while the camera turns.} Our subject error stays near its still-camera
level of 0.06 (0.04 on the half turn, 0.07 on the full), while TTM's rises to 0.09--0.12: a run is worse
than the same method's still-camera run in 8/28 cases for ours, against 20/28 for both TTM recipes.
Ours is lower than TTM in most sequences in all four paired comparisons (\cref{tab:supp_subject_pan_paired}).
These are the comparisons behind the main text's central subject-control claim
(\cref{sec:exp:subject}): our model follows a camera command and a subject command at the same time.

\textbf{Both parts of the recipe are needed.} Without the subject branch, the region transplant alone rises to
TTM's level (0.11 / 0.10) and is worse than still in 19/28 runs. Without the region transplant, the branch alone is
also at TTM's level (0.14 / 0.11) and misses the commanded turn (reach 1.20 on the half turn, 0.83 on the full).
With neither, the subject is not followed at all (0.16 / 0.21).
This confirms the main text's account (\cref{sec:exp:subject}): without the subject branch the subject drifts to
TTM's level, and without the region transplant the camera misses the commanded turn.

\textbf{Smoother motion.} Ours also moves more smoothly than TTM on Wan 2.2: VBench motion smoothness is
$98.9\%$ on the half turn and $98.5\%$ on the full turn, against $98.6\%$ / $98.1\%$ for TTM's camera recipe and
$98.7\%$ / $98.3\%$ for its cut-and-drag recipe (ours better on 11--14 of 14 sequences, $p{\le}0.03$).

\begin{table}[t]
  \centering\footnotesize
  \setlength{\tabcolsep}{2.5pt}
  \resizebox{\linewidth}{!}{%
  \begin{tabular}{@{}lcccc@{}}
    \toprule
    & Err.\,$\downarrow$ & Held\,$\uparrow$ & \multicolumn{2}{c}{Copy PSNR (dB)\,$\downarrow$}\\
    \cmidrule(l){4-5}
    Levels written & own path & stay & subject & hole\\
    \midrule
    \rowcolor{refgrey} None (camera only, floor) & 0.43 & 21/90 & 10.9 & 11.9\\
    \textbf{(A) 1--3, band} & 0.27 & 60/90 & \textbf{12.0} & \textbf{14.4}\\
    (B) 1--13                       & \textbf{0.04} & \textbf{81/90} & 20.9 & 29.6\\
    (C) 4--13, finer & 0.13 & 37/90 & 18.7 & 25.1\\
    \midrule
    \rowcolor{refgrey} Ours: reg.\ transplant + subj.\ branch & 0.15 & 61/90 & 12.7 & 15.7\\
    \bottomrule
  \end{tabular}}
  \caption{\textbf{Writing the finer levels (B, C) lowers the path error only by pasting the puppet and its
  inpainted hole; writing the control band (A) steers the subject without pasting.} Region transplant written
  into the listed levels, subject branch off; DAVIS, still camera; medians over the 80 sequences with a moving
  subject (90 for stay). Held: the subject stays within $2\%$ of the frame width under stay. Copy PSNR: to the
  puppet video, at the subject's commanded position and in the hole it left (higher = pasted). Grey rows, for
  reference: our camera model without a transplant (floor) and our region transplant with the
  subject branch.}
  \label{tab:supp_subject_placement}
\end{table}

\begin{figure*}[t]
  \centering
  \includegraphics[width=\linewidth]{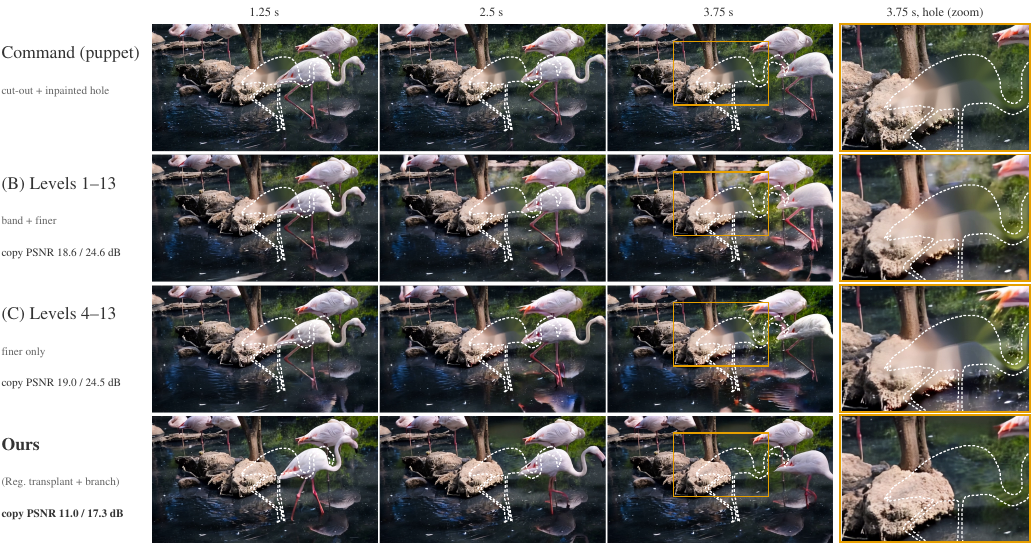}
  \caption{\textbf{Placement of the subject transplant, one example.} DAVIS~\cite{davis2017} \emph{flamingo},
  own path, still camera. White dashes: the flamingo at frame 0, which the puppet cuts out and fills by
  inpainting; right: zoom on that hole at 3.75\,s (orange box). Writing the puppet into the finer levels, with
  (B) or without (C) the control band, pastes the blurred inpainting onto the rock and the background; ours
  writes only levels 0--3 and the model redraws the rock and the background behind the flamingo.
  Copy PSNR: to the puppet video on this sequence (subject / hole zone; higher = pasted).}
  \label{fig:supp_subject_placement}
\end{figure*}

\subsection{Does placing the subject transplant in the control band matter?}
\label{sec:supp_subject_placement}

The main text writes the subject transplant into the coarse levels and lets the finer levels draw the
subject's appearance, the hole it leaves and the background (\cref{sec:method:recipe}).
We test this placement with the same region transplant written into three sets of levels:
(A) levels 1--3, the control band; (B) levels 1--13; (C) levels 4--13, the finer levels only.
The arms differ only in the levels written, and the subject branch is off in all three.
Our region transplant with the subject branch (levels 0--3) is shown for reference.
We run all 80 DAVIS~\cite{davis2017} sequences with a moving subject (own path) and all 90 for stay, with a
still camera.
Writing the finer levels can paint the puppet's pixels into the video, which motion measures alone would
reward, so every arm also gets a \emph{copy score}: the PSNR between the generated and the puppet video in the
subject's commanded position and in the inpainted hole it left behind (higher = more pasted).
Our camera model without any transplant sets the floor.

\textbf{Writing the finer levels pastes the puppet.} B and C follow the commanded path more closely than A
(error 0.04 and 0.13 against 0.27), but they do it by copying the puppet: their copy scores on the subject are
20.9 and 18.7\,dB, against 12.0 for A, 12.7 for our region transplant with the subject branch and 10.9
for the floor
(\cref{tab:supp_subject_placement}).
Written into the control band, the transplant avoids pasting the puppet: it steers the subject and leaves its
drawing to the model, as the main text states (\cref{sec:exp:subject}).

\textbf{Writing only the band mitigates the inpainted hole.} B and C also copy the hole the subject leaves
behind (copy scores 29.6 and 25.1\,dB, against 14.4 for A, 15.7 for ours and 11.9 for the floor).
In \cref{fig:supp_subject_placement}, they paste the blurred inpainting onto the rock and the background,
while ours redraws them behind the flamingo.
This is the main text's point that writing the transplant only into the control band mitigates inpainted holes
(\cref{sec:exp:subject}).

\textbf{Without the band, a subject cannot be held still.} Under stay, C holds only 37/90 subjects still,
against 60/90 for A (and 61/90 for ours).
When the command disagrees with the motion the model has decided at the coarse levels, writing the finer
levels does not override it.
This is the subject's counterpart of camera arm C (main text, \cref{tab:placement}) and supports the main
text's finding that motion is decided at levels 1--3 (\cref{sec:method:control}).

\section{Revisit Scene Experiments}
\label{sec:supp_memory}

\subsection{Memory against models trained for memory}
\label{sec:supp_mem_baselines}
\begin{table*}[t]
  \centering\small
  \setlength{\tabcolsep}{3.2pt}
  \begin{tabular}{@{}llcccccc@{}}
    \toprule
    System & Memory & Return PSNR $\uparrow$ & LPIPS $\downarrow$ & Gain (dB) $\uparrow$ & Return err.\ $\downarrow$ & Turns & Min.\\
    \midrule
    \rowcolor{refgrey} Ours & none & 12.9 / 12.8 & 0.510 & -- & 7.5 / 11.7$^\circ$ & 8/8 & 3.8\\
    Ours & 88\,KB & 21.4 / 24.8 & 0.237 & +10.2 {\footnotesize[9.3, 11.2]} & 5.2 / 3.1$^\circ$ & 8/8 & 4.0\\
    \textbf{Ours} & 88\,KB + redraw & 22.8 / 23.6 & 0.151 & +10.3 {\footnotesize[9.0, 11.6]} & 0.6 / 0.7$^\circ$ & 8/8 & 5.9\\
    Ours, Interact base & none & 13.6 / 12.2 & 0.486 & +0.1 {\footnotesize[$-$1.2, 1.2]} & 7.0 / 5.0$^\circ$ & 8/8 & 3.9\\
    \textbf{Ours, Interact base} & 84\,KB + redraw & \textbf{26.8 / 27.9} & \textbf{0.119} & \textbf{+14.5} {\footnotesize[12.7, 16.1]} & 0.8 / \textbf{0.5}$^\circ$ & 8/8 & 6.2\\
    \midrule
    Interact, neutral prompt & last clip & 10.7 / 13.2 & 0.631 & $-$0.9 {\footnotesize[$-$2.1, 0.4]} & 49.6 / 26.6$^\circ$ & 0/8 & 4.1\\
    Interact, motion prompts & last clip & 10.9 / 11.7 & 0.610 & $-$1.6 {\footnotesize[$-$2.4, $-$0.8]} & 28.4 / 16.5$^\circ$ & 4/8 & 4.1\\
    Spatia & point cloud & 18.4 / 17.9 & 0.262 & -- & 1.6 / 2.2$^\circ$ & 7/8 & 47.5\\
    UCM & on & 21.8 / 22.9 & 0.192 & -- & \textbf{0.3} / 1.0$^\circ$ & 8/8 & 14.1\\
    \bottomrule
  \end{tabular}
  \caption{\textbf{Our store brings both bases back to the start view ($+10.3$ and $+14.5$\,dB over no
  memory); on the Interact base it gives the highest return PSNR and lowest LPIPS of all systems.} The 20\,s
  chain in full. Return PSNR (dB) and return error on
  the first / second return; LPIPS averaged over both. Gain: mean gain in return PSNR over our
  no-memory chain, with a 95\% bootstrap interval, for systems that share our first call. Min.: minutes to generate
  the 20\,s video on one A100 (see Generation time).}
  \label{tab:supp_mem_baselines}
\end{table*}

This section details \cref{tab:mem_baselines,fig:mem_baselines} of the main text (\cref{sec:exp:memory});
\cref{tab:supp_mem_baselines} gives the 20\,s chain in full.
We encourage readers to view the 20\,s chains on the \projectpage{} (``Return memory'' tab), ours on
the Interact base next to UCM, Spatia and Interact.

\paragraph{Chain.}
The 20\,s chain of the main text: 8 held-out RealEstate10K~\cite{re10k} scenes, four 5\,s legs (pan right
$45^\circ$, back, pan left $45^\circ$, back), 321 frames at 16\,fps, and one neutral prompt (``A video clip showing
smooth camera motion through a real-world scene.'').

\paragraph{Scoring.}
Each return leg is scored against the outward leg before it, played backwards, i.e.\ frame by frame at the same
camera pose: PSNR and LPIPS per latent frame, the first excluded.
The camera's yaw on each leg is estimated from RAFT~\cite{raft} flow with a rotation homography; the return error is
$|\text{yaw(out)}+\text{yaw(back)}|$, and \emph{turns} counts the scenes in which all four legs turn the commanded way.

\paragraph{Baselines.}
All baselines run as released, with their own default settings.
\emph{UCM}~\cite{ucm}: memory on, one 321-frame call at $352{\times}640$.
\emph{Spatia}~\cite{spatia}: 481 frames at 24\,fps ($121+5{\times}72$, its generation rounds) at $1248{\times}704$, on
the same camera path resampled to 24\,fps; its video is resampled to 321 frames before scoring. It has no memory-off
mode.
\emph{InfinityStar-Interact}~\cite{infinitystar}: the released model continues a clip from a spatially downsampled
summary of the previous one plus its last two frames, and has no camera input. It starts from our first 5\,s call and
adds three 5\,s continuations, with the neutral prompt or with one prompt per leg that names its motion.
UCM and Spatia generate their own first 5\,s from the input frame; all other systems share our first call.

\paragraph{Ours.}
\emph{Ours}: the frozen base model with our camera adapter (\cref{sec:method}) and the 88\,KB store of levels 0--10
with redrawn detail (\cref{sec:exp:memory}).
\emph{Ours, Interact base}: the released Interact model, frozen, with our camera adapter retrained in Interact's
clip-to-clip mode (initialised from ours; 4,800 steps on 4,902 10\,s DL3DV~\cite{dl3dv} clips, 2 A100s), and the same
store and redraw (84\,KB).

\paragraph{Ours returns closer than UCM and Spatia.}
Paired over the 8 scenes (mean of both returns, 95\% bootstrap intervals), our return PSNR is $0.8$\,dB $[0.3, 1.4]$
above UCM's and $5.0$\,dB $[4.4, 5.8]$ above Spatia's. On the Interact base it is $5.0$\,dB $[4.2, 5.8]$ above UCM's
and $9.2$\,dB $[8.3, 10.0]$ above Spatia's, in 8/8 scenes each.
This is the main text's result that our training-free memory returns as accurately as models trained for memory
(\cref{sec:exp:memory}).

\paragraph{The gain comes from the store.}
Without memory, the Interact base with our adapter returns no better than our base ($+0.1$\,dB), and Interact's
own carry-over of the last clip does no better either.
Its $+14.5$\,dB therefore comes from the store: as the main text states, our store carries over to a second
next-scale model (\cref{sec:exp:memory}).

\paragraph{Why ours returns closer.}
UCM and Spatia remember pixels. To show a stored view again they must place it in the new view through estimated
depth: UCM warps its positional encodings with per-frame depth, Spatia projects a point cloud and re-renders it. Depth
errors, disocclusion holes and colour drift can then enter each return; in Spatia they compound: its
second return is 0.5\,dB below its first and its colours wash out.
Our store keeps the model's own coarse tokens, the levels at which the hierarchy decides layout (\cref{sec:method}).
Re-ordered along the return path and forced, they let the same generator redraw the same scene from the top of its
hierarchy; nothing is reprojected, so no geometry has to be right.

\begin{figure*}[t]
  \centering
  \includegraphics[width=\linewidth]{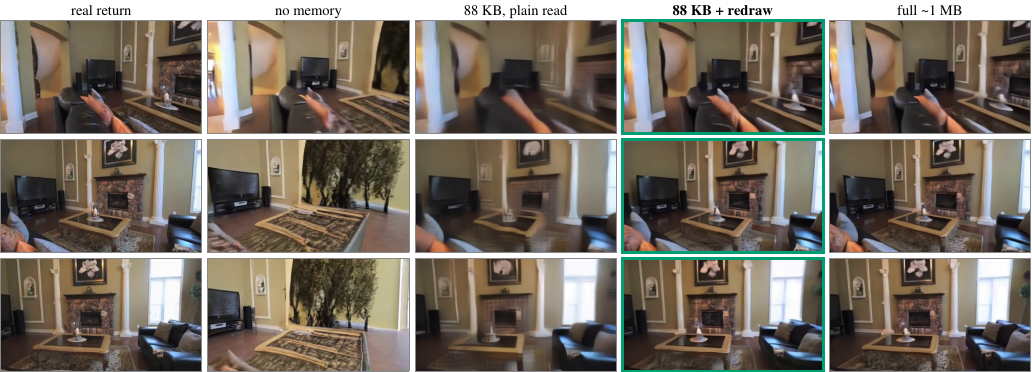}
  \caption{\textbf{Turning back in a living room}, three frames of the return. Held-out RealEstate10K~\cite{re10k}
  return. Without memory, the model invents a new room. The 88\,KB coarse store restores the room but is blurry;
  redrawing the detail sharpens it (25.0 vs.\ 22.5\,dB, LPIPS 0.23 vs.\ 0.39), close to a ${\sim}1$\,MB store
  (26.2\,dB).}
  \label{fig:supp_memory}
\end{figure*}

\paragraph{One Spatia failure.}
In scene \texttt{001573921974bc70}, an aerial view, Spatia's video washes out into haze from about frame 300 of 481;
its third leg turns the wrong way ($+51^\circ$) and its return LPIPS is 0.60/0.71. Without this scene Spatia returns
at 18.8\,dB (LPIPS 0.206), against 23.5 for ours and 27.8 for ours on the Interact base; the ranking is unchanged.

\paragraph{Generation time.}
One A100 80GB, model loading excluded, as for the other costs in this paper.
UCM: 842--852\,s per chain over the 8 scenes (14.1\,min). Spatia: median 2,852\,s (2,726--3,730\,s; 47.5\,min).
The InfinityStar-based chains were timed on one idle A100.
Ours: the 20\,s chain of four 5\,s calls takes 3.8\,min without memory and 4.0 with the
88\,KB reader; redraw regenerates the stored clip at each of the two returns (5.9\,min).
Ours on the Interact base: 3.9\,min without memory and 6.2 with redraw. Interact as released: 4.1\,min
(our first call plus its three continuations; the motion-prompt variant only changes the text).
With redraw, ours generates the 20\,s chain $2.4\times$ faster than UCM and $8\times$ faster than Spatia (main text,
\cref{tab:mem_baselines}): a store read by the generator itself needs no memory model and no 3D reconstruction.

\subsection{Returns that do not retrace the stored clip}
\label{sec:supp_mem_revisits}

The main text states that the same reader handles returns that do not retrace the stored clip, gaining 11--14\,dB
in every case (\cref{sec:exp:memory}). \Cref{tab:supp_mem_revisits} gives the five cases, each on 8 held-out
RealEstate10K scenes with our camera model and the 88\,KB store with redrawn detail, in chains of 15--25\,s: a return
that overlaps the stored clip only partly, one that spans two stored clips, one twice as fast, a tilted return, and
five $90^\circ$ pans in one direction, whose fifth call sees the first view again.

\begin{table}[t]
  \centering\footnotesize
  \setlength{\tabcolsep}{3pt}
  \begin{tabular}{@{}lccc@{}}
    \toprule
    Revisit & No memory & Ours & Gain (dB)\\
    \midrule
    Partial overlap with the stored clip & 13.1 & 24.7 & $+11.6$\\
    Return over two stored clips          & 12.9 & 25.6 & $+12.7$\\
    $2\times$ faster return               & 12.1 & 23.5 & $+11.4$\\
    Tilted return                         & 12.9 & 27.2 & $+14.3$\\
    Loop closed in the same direction     & 9.8  & 23.9 & $+14.1$\\
    \bottomrule
  \end{tabular}
  \caption{\textbf{Every return that does not retrace the stored clip gains 11--14\,dB}, as stated in the main text.
  Revisit PSNR (dB) against the model's own earlier frames at the same pose; 8 held-out RealEstate10K scenes each;
  ours: our camera model with the 88\,KB store and redrawn detail.}
  \label{tab:supp_mem_revisits}
\end{table}

\paragraph{Close to a store $11.6\times$ larger.}
Main-text \cref{tab:mem_phase1} scores the first three cases with the reader before redraw, which gains about
9\,dB. Redrawing the detail adds the rest and brings the 88\,KB store close to the ${\sim}1$\,MB store of all levels
there: 24.7 vs.\ 24.9\,dB on the partial overlap, 25.6 vs.\ 26.3 over two stored clips, and 23.5 vs.\ 23.3 on the
$2\times$ faster return. As on the real returns of the main text, redrawing the detail closes most of the gap to the
full store at the same 88\,KB (\cref{sec:exp:memory}).

\subsection{What to store and how to read it}
\label{sec:supp_mem_store}

\paragraph{Keep all of levels 0--10.}
Keeping all of levels 0--10 beats stopping the store at level 7, 5 or 3 in all 16 revisits we tested (4 scenes
$\times$ 4 revisit types), at a cost of $3.1$, $5.8$ and $8.2$\,dB. This is the main text's finding that keeping fewer levels loses in every revisit we tested, and it follows
the layout band, which ends at level 10 (\cref{sec:method:render,sec:exp:memory}).

\paragraph{Redraw the detail.}
Redrawing levels 11--13 before re-encoding closes $61\%$ of the PSNR gap and $79\%$ of the LPIPS gap to the
${\sim}1$\,MB store on the real returns, at the same 88\,KB (main text, \cref{sec:exp:memory});
\cref{fig:supp_memory} shows one return. In the 20\,s chains it lowers LPIPS from $0.24$ to $0.15$ and the return
error from $5.2/3.1^\circ$ to $0.6/0.7^\circ$ (\cref{tab:supp_mem_baselines}).
This is the reader of the main text (\cref{sec:method:recipe}): the stored coarse tokens fix the layout, and the
model redraws the detail.

\subsection{Other camera motions}
\label{sec:supp_mem_long}

Ours on the Interact base (\cref{tab:supp_mem_baselines}) runs the same 84\,KB store with redraw on further 20\,s
trajectories from held-out RealEstate10K scenes (\cref{tab:supp_mem_motions}): larger pans, a sideways truck, and a
$90^\circ$ return in one clip that reads two stored clips at double speed. In every case the frame at 20\,s comes back
to the start view, at 22.9--24.9\,dB against 10.5--11.8\,dB without memory.

\begin{table}[t]
  \centering\footnotesize
  \setlength{\tabcolsep}{3pt}
  \begin{tabular}{@{}lccc@{}}
    \toprule
    Trajectory (four 5\,s calls) & Scenes & No memory & Memory\\
    \midrule
    Pan $45^\circ$ right, back, left, back   & 16 & 11.8 & 24.4\\
    Pan $90^\circ$ right, back, left, back   & 16 & 11.2 & 22.9\\
    Truck right, back, left, back            & 8  & 10.5 & 23.1\\
    Right $45^\circ$, hold, right $45^\circ$, back $90^\circ$ & 8 & 11.0 & 24.9\\
    \bottomrule
  \end{tabular}
  \caption{\textbf{Memory brings every trajectory back to the start view: 22.9--24.9\,dB at 20\,s, against
  10.5--11.8\,dB without memory.} Other camera motions over 20\,s. Ours on the Interact base, 84\,KB store with redraw. PSNR
  (dB) of the frame at 20\,s against the start view (one frame, not the whole return as in \cref{tab:supp_mem_baselines}).}
  \label{tab:supp_mem_motions}
\end{table}

\section{Scope and Future Work}
\label{sec:limitations}

We identify the hierarchy and build our controls on InfinityStar~\cite{infinitystar}, to our knowledge the only released discrete 
next-scale model for open-domain video, and its Interact variant. On RAYNOVA, the closest released model, our hierarchy
analyses carry over (\cref{sec:supp_other_discrete}), but its weak camera following confounds them.
The band boundaries may differ on other models, but the training-free transplant protocol can locate them on any
next-scale model that follows the camera closely enough for the transfer to be measured.

To demonstrate an added advantage of identifying the token hierarchy,
we show coarse levels acting as memory with camera pose and revisiting static scenes. 
Returning to a scene whose content has changed is left for future work.
Image quality is set by the base model and its tokenizer, not by our store or controls, so a sharper next-scale
model would raise it without changing them.

{
    \small
    \bibliographystyle{ieeenat_fullname}
    \bibliography{main}
}

\end{document}